\documentclass{article}

\usepackage{microtype}
\usepackage{graphicx}
\usepackage{amsfonts,amsmath,mathtools,bm}
\usepackage{booktabs} 

\usepackage{hyperref}

\usepackage[accepted]{icml2021}

\usepackage{microtype}
\usepackage{graphicx}
\usepackage{booktabs} 
\usepackage{url}            
\usepackage{amsfonts}       
\usepackage{nicefrac}       
\usepackage{xspace}
\usepackage{amsmath}
\usepackage{amssymb}
\usepackage{float}
\usepackage{lipsum}
\usepackage{bigcenter}
\usepackage{newtxtext,amsthm,bbm}
\usepackage{dsfont}
\usepackage{tikz,soul}
\usetikzlibrary{backgrounds,decorations.pathreplacing,intersections}
\usepackage{mathtools}
\usepackage{thmtools}
\usepackage{thm-restate}

\declaretheorem[name=Lemma]{lemma}

\newtheorem{definition}{Definition}
\newtheorem{assumption}{Assumption}
\newtheorem{remark}{Remark}

\newtheorem{problem}{Problem}
\newtheorem{example}{Example}
\usetikzlibrary{shapes,shadows,automata,arrows,positioning,calc}
\usepackage{natbib}
\usepackage[utf8]{inputenc} 
\usepackage[T1]{fontenc}    
\usepackage{xargs}       
\usepackage{enumitem}
\usepackage[scr=boondox]{mathalfa}
\usepackage[colorinlistoftodos,prependcaption,textsize=tiny]{todonotes}
\usepackage{algpseudocode}
\usepackage{subcaption}
\DeclareCaptionSubType*{algorithm}

\DeclareCaptionLabelFormat{alglabel}{Algorithm~#2}

\newcommandx{\rlm}[2][1=]{\todo[linecolor=violet,backgroundcolor=violet!25,bordercolor=violet,#1]{\textbf{Romain:} #2}}
\newcommandx{\rfm}[2][1=]{\todo[linecolor=red,backgroundcolor=red!25,bordercolor=red,#1]{\textbf{Raphael:} #2}}

\newcommand*{\mytop}{\mathrel{\scalebox{0.5}{$\top$}}}
\newcommand*{\mybot}{\mathrel{\scalebox{0.5}{$\bot$}}}
\newcommand*{\mydiese}{\mathrel{\scalebox{0.5}{$\#$}}}
\newcommand*{\myplus}{\mathrel{\scalebox{0.5}{$+$}}}
\newcommand*{\myminus}{\mathrel{\scalebox{0.5}{$-$}}}

\DeclareMathOperator*{\argmax}{argmax}
\DeclareMathOperator*{\sign}{sign}

\icmltitlerunning{Batched Bandits with Crowd Externalities}

\begin{document}

\twocolumn[
\icmltitle{Batched Bandits with Crowd Externalities}




\begin{icmlauthorlist}
\icmlauthor{Romain Laroche}{msr} 
\icmlauthor{Othmane Safsafi}{pmc} 
\icmlauthor{Rapha\"el F\'eraud}{org} 
\icmlauthor{Nicolas Broutin}{pmc} 
\end{icmlauthorlist}

\icmlaffiliation{msr}{Microsoft Research, Montr\'eal, Quebec, Canada}
\icmlaffiliation{pmc}{Sorbonne Universit\'e, Paris, France}
\icmlaffiliation{org}{Orange Labs, Lannion, France}

\icmlcorrespondingauthor{Romain Laroche}{rolaroch@microsoft.com}

\icmlkeywords{Emergent Language, Agent-based, Iterative Learning, Reinforcement Learning, Emergent behavior, Learning Phases}

\vskip 0.3in
]



\printAffiliationsAndNotice{}  


\renewcommand*{\theenumi}{\thesection.\arabic{enumi}}

\begin{abstract}
    In Batched Multi-Armed Bandits (BMAB), the policy is not allowed to be updated at each time step. Usually, the setting asserts a maximum number of allowed policy updates and the algorithm schedules them so that to minimize the expected regret. In this paper, we describe a novel setting for BMAB, with the following twist: the timing of the policy update is not controlled by the BMAB algorithm, but instead the amount of data received during each batch, called \textit{crowd}, is influenced by the past selection of arms. We first design a near-optimal policy with approximate knowledge of the parameters that we prove to have a regret in $\mathcal{O}(\sqrt{\frac{\ln x}{x}}+\epsilon)$ where $x$ is the size of the crowd and $\epsilon$ is the parameter error. Next, we implement a UCB-inspired algorithm that guarantees an additional regret in $\mathcal{O}\left(\max(K\ln T,\sqrt{T\ln T})\right)$, where $K$ is the number of arms and $T$ is the horizon.
\end{abstract}

\renewcommand{\topfraction}{0.8}
\renewcommand{\textfraction}{0.1}
\section{Introduction}
This paper tackles a novel instance of Batched Multi-Armed Bandits~\citep[BMAB,][]{Perchet2016,Gao2019}, where the timing of updates is constrained by the environment, but the \textit{crowd}, \textit{i.e.} the number of samples collected in the next batch depends on the arms that have been pulled in past batches. While we believe that there are many more applications to this setting (see the broader impact section for some of them), we will use the following application example to illustrate and motivate our work.
\begin{example}[Service in production] The service may only be updated once everyday over night. We have two (or more) options to deliver the service:
    \renewcommand{\labelenumi}{Arm \arabic{enumi}.}
    \begin{enumerate}[leftmargin=1.5cm]
        \item with advertisement: it yields income but low user satisfaction,
        \item without advertisement: it yields costs but high user satisfaction.
    \end{enumerate}
    \label{ex:service}
\end{example}
Playing the first arm is profitable but decreases the crowd, and playing the second arm increases the crowd but is loss-making. In practice, the interplay between the users, their task success, the crowd dynamics may be extremely complex. While strongly motivated by real-world scenarios, in order to control the complexity of our study, which is the first of its kind, we will consider an idealized setting by making the following series of assumptions:
\begin{assumption}[Idealized setting assumptions]
    \leavevmode
    \renewcommand{\labelenumi}{\emph{A1(\roman{enumi})}}
    \begin{enumerate}[leftmargin=1.1cm]
        \item The crowd size at next time step is the sum of individual growth: the number of samples to be collected at the next round induced by each arm pull. This sum is then capped by a known full capacity.
        \item Individual growths and rewards are independent, identically distributed, and observable (even when the crowd has been capped).
        \item Crowd size at time $t$ is known beforehand.
    \end{enumerate}
    \label{ass:stylized}
\end{assumption}

Some of these assumptions could have been worked around, but we decided not to for clarity reasons, in order to remain in a pristine setting that is already sufficiently complex by itself. Under Assumptions \ref{ass:stylized}, it may happen that the service is not sustainable: it is impossible to gain money while maintaining the crowd. In this case, the objective is to make the most of the initial crowd. In the other case, we will show that the optimal policy is to first invest to grow the crowd until reaching its full capacity and then to collect the return on investment while maintaining the full crowd.


While the environment is naturally a Markov Decision Process (MDP), we cannot use classic Reinforcement Learning (RL) algorithms to solve our setting, because only one trajectory is allowed, and exploration would lead it to the terminal state where no user remains in the crowd. Thus, our global objective is to design a bandit algorithm~\citep{Bubeck2012} that deals with the exploration/exploitation trade-off when the crowd dynamics and the rewards are unknown. The exploration intends to reduce the model error. The exploitation intends to yield high rewards. The difficulty of our setting is that this trade-off must be performed under a survival effort: make sure that the crowd runs out only when the model is known to be unsustainable with high probability.

The effect of arm pulls on future rewards has been extensively studied in previous works.
In restless bandit \citep{Whittle1988}, every time an arm is sampled, its state changes according to a transition matrix $q$, while otherwise, its state changes according to another transition matrix $\tilde{q}$. It has been showed that computing the optimal policy of restless bandits is PSPACE-hard \citep{Papadimitriou99}, and hence relaxation techniques are used for finding an approximation \citep{Whittle1988,Guha2010}. There exists a lot of variations of the standard multi-armed bandits, where the future rewards depends on the played actions and where finding the optimal policy is not intractable. In mortal bandits \citep{Deepayan2009}, each arm has a lifetime after which it disappears. In scratch games \citep{Feraud2013}, an urn model is used for handling the lifetime of arms. In Multi-Armed Bandits with known trend \citep{Bouneffouf2016}, the future reward distribution depends on a known function of the number of times the arm has been sampled, while in recovering bandits the trend function is learnt \citep{Pike2019}. Unlike this line of works, we consider here that sampling an arm does not modify the future rewards, but changes the arrival process of new subjects. To the best of our knowledge, the only work that considers the impact of the sampled actions to the arrival process of new subjects is bandit with positive externalities \citep{Shah2018}. While in the proposed problem setting the pull of arms influences the number of subjects that will arrive at the next time periods, in \cite{Shah2018}, the pull of arms influences the type of subjects that will arrive at the next time periods. Externalities have also been widely studied in economics~\citep{Cornes1996,Klenow2005}.

The contributions of this article are the following: Section \ref{sec:problemstatement} formalizes the problem and casts it as a Markov Decision Process (MDP). Expressed this way, the policy optimization is intractable because of the stochasticity in the environment. We search for an approximate solution by solving the deterministic Reduced On-expectation MDP (ROeMDP). Section \ref{sec:analysis} develops the theory and proves the near optimality of the ROeMDP solution in $\mathcal{O}(\sqrt{\frac{\ln x}{x}}+\epsilon)$, where $x$ is the current crowd size, and $\epsilon$ is the error on the problem parameters (Theorem \ref{th:optimalPMDPerror}). Building on these findings, Section \ref{sec:bandit} introduces a novel UCB algorithm for the problem and proves it to have an additional regret in $\mathcal{O}\left(\max(K\ln T,\sqrt{T\ln T})\right)$, where $K$ is the number of arms and $T$ is the horizon, as compared to the ROeMDP approximate solution (Theorem \ref{th:bandittheorem}). Section \ref{sec:experiments} runs some numerical simulations to validate the theoretical findings. We observe the unexpected result that the bandit algorithm often outperforms the ROeMDP solution with the true parameters. This phenomenon is explained by the bias induced by UCB's optimism. Section \ref{sec:conclusion} concludes the main document with perspectives for future work. Supplementary material includes all proofs and an overview of the broader impacts.

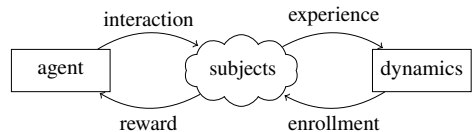
\begin{figure}[t]
	\centering
	\newcommand*{\xMin}{0}
	\newcommand*{\xMax}{5}
	\newcommand*{\yMin}{0}
	\newcommand*{\yMax}{5}	
	\newcommand*{\tikzfigscale}{0.8}
	\begin{tikzpicture}[scale=\tikzfigscale, every node/.style={scale=\tikzfigscale}, ->, shorten >=1pt,auto,node distance=3cm,very thin]
        \tikzstyle{state}=[circle, draw, fill=white,draw=black,text=black, text width = 1em, text centered]
        \tikzstyle{sensor}=[circle, draw, fill=white,draw=black,text=black, text width = 1em, text centered]
        \tikzstyle{device}=[rectangle, draw, fill=white,draw=black,text=black, text width=4em, text centered, minimum height=2em]
        \tikzstyle{network}=[cloud, draw,cloud puffs=10,cloud puff arc=120, aspect=2, inner ysep=0.3em, text width = 3em, minimum height =2em, text centered]
        
        \node[device] (S) {agent};
        \node[network] (U) [right of=S] {subjects};
        \node[device] (D) [right of=U] {dynamics};
        \path (S) edge[bend left] node{interaction} (U);
        \path (U) edge[bend left] node{reward} (S);
        \path (U) edge[bend left] node{experience} (D);
        \path (D) edge[bend left] node{enrollment} (U);
    \end{tikzpicture}
	\caption{Multi-batch setting.}
	\vspace{-20pt}
	\label{fig:benchmark}
\end{figure}
\section{Problem formalization}
\label{sec:problemstatement}
\subsection{Problem statement}
In this paper we study a setting illustrated on Figure \ref{fig:benchmark}, where, at each time step $t$, an agent independently and identically interacts with a set of subjects, called the crowd, which size is denoted by $x_t$. For each subject, the agent selects its play among $K$ arms and receives a reward as a result, similarly to what happens in any stochastic multi-armed bandit (MAB) setting. But contrary to standard MABs, we consider that the samples are received by batches, and that the number of subjects in the next batch depends on the past pulls. The agent decides an action $a_t$: the number of pulls on each arm, indifferently spread among subjects. Each pull on Arm $k$ triggers an interaction $\tau$ yielding a reward $\dot{r}_\tau$ sampled from distribution $r_k$, and a growth $\dot{g}_\tau$ sampled from distribution $g_k$. $\dot{g}_\tau \in \mathbb{N}$ is the number of subjects being enrolled for next time step $t+1$ stemming from interaction $\tau$. The goal is therefore to optimize the selection of arms, accounting both for the immediate global reward $r_t= \sum_{\tau=1}^{x_t} \dot{r}_\tau$ and the future ones that are directly depending on the global crowd $x_{t+1}=\min(\sum_{\tau=1}^{x_t} \dot{g}_\tau, x_{\mytop})$, where $x_{\mytop}$ is the known maximal population. We assume that $\dot{r}_\tau$, and $\dot{g}_\tau$ are observed for each arm pull (Assumption A1(ii)). 

In contrast with \citet{Shah2018} and \citet{Laroche2018AS}, the externality of our setting is simpler, since it amounts to a factor effect over the rewards that is the same for all arms. However, it may be used as a controllable feature, and as such, may be regarded as a multi-state problem and therefore a Reinforcement Learning task~\citep[RL,][]{Sutton1998}. Still, contrary to classic RL tasks, the decision process involves a single trajectory with terminal states.

Notations: Let $\Delta_{S}$ denote a distribution over set $S$. Let $[n]$ be the set of integers $1\leq i\leq n$. We write $\overline{g}_k=\mathbb{E}g_k$, which we call the expected growth of arm $k$. $\overline{r}_k=\mathbb{E}r_k$ is similarly defined as its expected reward. $\overline{g}_{\mybot}$ (resp. $\overline{g}_{\mytop}$) is the minimum (resp. maximum) expected growth over the arms: $\overline{g}_{\mybot} = \min_{k\in[K]} \overline{g}_k$ and $\overline{g}_{\mytop} = \max_{k\in[K]} \overline{g}_k$, and the maximum expected reward is denoted by $\overline{r}_{\mytop} = \max_{k\in[K]} \overline{r}_k$.

We formalize the problem we intend to solve as follows.

\begin{problem}
    Design and analyze an algorithm $\mathfrak{A}$, that, at each time step $t$, takes as a argument the history of past experience and returns an action $a_t$, in order to maximize the following $\gamma$-discounted objective\footnote{The discount is used to prevent infinite returns. We allow ourselves to choose it as close to 1 as needed.}:
		\begin{align}
		    &V_{\mathfrak{A}}(x_0) = \mathbb{E}\left[\sum_{t=0}^\infty \gamma^t \sum_{\tau=1}^{x_{t}} \dot{r}_\tau \right], \\
		    &\text{with }\!\!\left\{ 
		    \begin{array}{l}
		         \!\!\! a_t = \mathfrak{A}(h_t), \quad k_\tau \in a_t, \quad \dot{r}_\tau \sim r_{k_\tau}, \quad \dot{g}_\tau \sim g_{k_\tau}, \\
		         \!\!\! h_{t+1} = h_t\cup(k_\tau,\dot{r}_\tau,\dot{g}_\tau), x_{t+1} = \min(\sum_{\tau=1}^{x_t} \dot{g}_\tau, x_{\mytop})
		    \end{array}\right.\nonumber
		\end{align} 
	and where history $h_0$ is initialized as $\emptyset$.
\end{problem}

\subsection{Model of the environment as MDPs}
The crowd, and therefore further rewards, depends on past actions. Hence, we need a Markov Decision Process to model the setting. We frame this type of dynamics as a Populated MDP (PMDP).
\begin{definition}[Populated MDP]
    A Populated MDP (PMDP) is a stochastic MDP $\langle \mathcal{X}_p, \mathcal{A}_p, P_p, R_p, \gamma \rangle$, where $\mathcal{X}_p = [x_{\mytop}]$ is the size of the population, $\mathcal{A}_p = [K]^{x_{\mytop}}$ is the action space, the stochastic transition function is $P_p(x,a) = \min\left\{\sum_{i=1}^{x} \dot{g}_{a[i]},x_{\mytop}\right\}$, the stochastic reward function is the sum of individual stochastic rewards $R_p(x,a) = \sum_{i=1}^{x} \dot{r}_{a[i]}$, and $\gamma$ is the discount factor.
    \label{def:PMDP}
\end{definition}

$V_p^\psi$ and $Q_p^\psi$ denote the values of a policy $\psi$ in the PMDP. We write the optimal values $V_p^*$ and $Q_p^*$, and $\psi^*$ may refer to any optimal policy in the PMDP. Expressed this way, the policy optimization is intractable because of the stochasticity in the environment. We are going to search for an approximate solution by solving the following deterministic MDP formulation, coined On-expectation MDP (OeMDP), which we define below:
\begin{definition}[On-expectation MDP]
    We define the On-expectation MDP (OeMDP) as the tuple $\langle \mathcal{X}_o, \mathcal{A}_o, P_o, R_o, \gamma \rangle$, where the state space is now continuous: $\mathcal{X}_o = [0,x_{\mytop}] \subset \mathbb{R}$, the action space is a distribution over the arms: $\mathcal{A}_o = \Delta_{[K]}$, the deterministic transition function is the expectation of growth: $P_o(x,a) = \min\left\{x\sum_{k\in[K]} a_k \overline{g}_k, x_{\mytop}\right\},$ and the deterministic reward function is the expectation of rewards: $R_o(x,a) = x\sum_{k\in[K]} a_k \overline{r}_k$.
    \label{def:OeMDP}
\end{definition}

We underline that the state space has to be defined on real numbers, since the expectation over a integer random variable lives in the real numbers. Consequently, the action space is a distribution over arms. $V_o^\psi$ and $Q_o^\psi$ denote the values of a policy $\psi$ in the OeMDP. We write the optimal values $V_o^{\mydiese}$ and $Q_o^{\mydiese}$, and $\psi^{\mydiese}$ refers to any optimal policy in the OeMDP.  In the OeMDP, any action $a\in\Delta_{[K]}$ has an effective growth $g_a=\sum_{k=1}^K a_k \overline{g}_k$. Conversely any value $g\in[\overline{g}_{\mybot}, \overline{g}_{\mytop}]$ is achievable by an interpolation between two arms, and once $g$ is selected, then an optimal policy must be only selecting actions that are maximizing the reward under the constraint of having a growth equal to $g$. We formalize this observation with the transformation operations $\Psi$ and $\mathcal{R}$ defined below.

\begin{definition}[Transformed action set and reward function]
    We define the transformed action set $\Psi$ and the transformed reward function $\mathcal{R}$ as follows:
    \begin{align}
        \left\{ 
		    \begin{array}{l}
		        \Psi(g) = \displaystyle\argmax_{a\in\Delta_{[K]}, \text{ s.t. } \sum_{k=1}^K a_k \overline{g}_k = g} \quad \sum_{k=1}^K a_k \overline{r}_k, \\
                \mathcal{R}(g) = \displaystyle\max_{a\in\Delta_{[K]}, \text{ s.t. } \sum_{k=1}^K a_k \overline{g}_k = g}\quad \sum_{k=1}^K a_k \overline{r}_k.
        \end{array}\right.\nonumber
	\end{align} 
    \label{def:transformation}
\end{definition}

Definition \ref{def:transformation} allows us to simplify the OeMDP formalization into a Reduced On-expectation MDP (ROeMDP), which is defined below.

\begin{definition}[Reduced On-expectation MDP]
    We define the following deterministic Reduced On-expectation MDP (ROeMDP) as $\langle \mathcal{X}_o, \mathcal{G}, P_r, R_r, \gamma \rangle$, where the actions are the growth $\mathcal{G} = [\overline{g}_{\mybot}, \overline{g}_{\mytop}] \subset \mathbb{R}$\footnote{For simplicity, we assume in all our proofs that $1\in\mathcal{G}$, but the lemmas and theorems still stand when this is not the case with minor changes in their formulations, and in their consequent proofs.}, the transition and reward functions are modified accordingly: $P_r(x,g) = \min\{xg,x_{\mytop}\}$ and $R_r(x,g) = x\mathcal{R}(g)$.
    \label{def:ROeMDP}
\end{definition}

Similarly to notations of PMDP and OeMDP, $V_r^\pi$ and $Q_r^\pi$ denote the values of a policy $\pi$ in the ROeMDP. We write the optimal values $V_r^{\mydiese}$ and $Q_r^{\mydiese}$, and $\pi^{\mydiese}$ refers to any optimal policy in the ROeMDP. 

\begin{figure*}[t]
    \bigcentering
    \includegraphics[trim = 15pt 5pt 35pt 25pt, clip, width=0.33\textwidth]{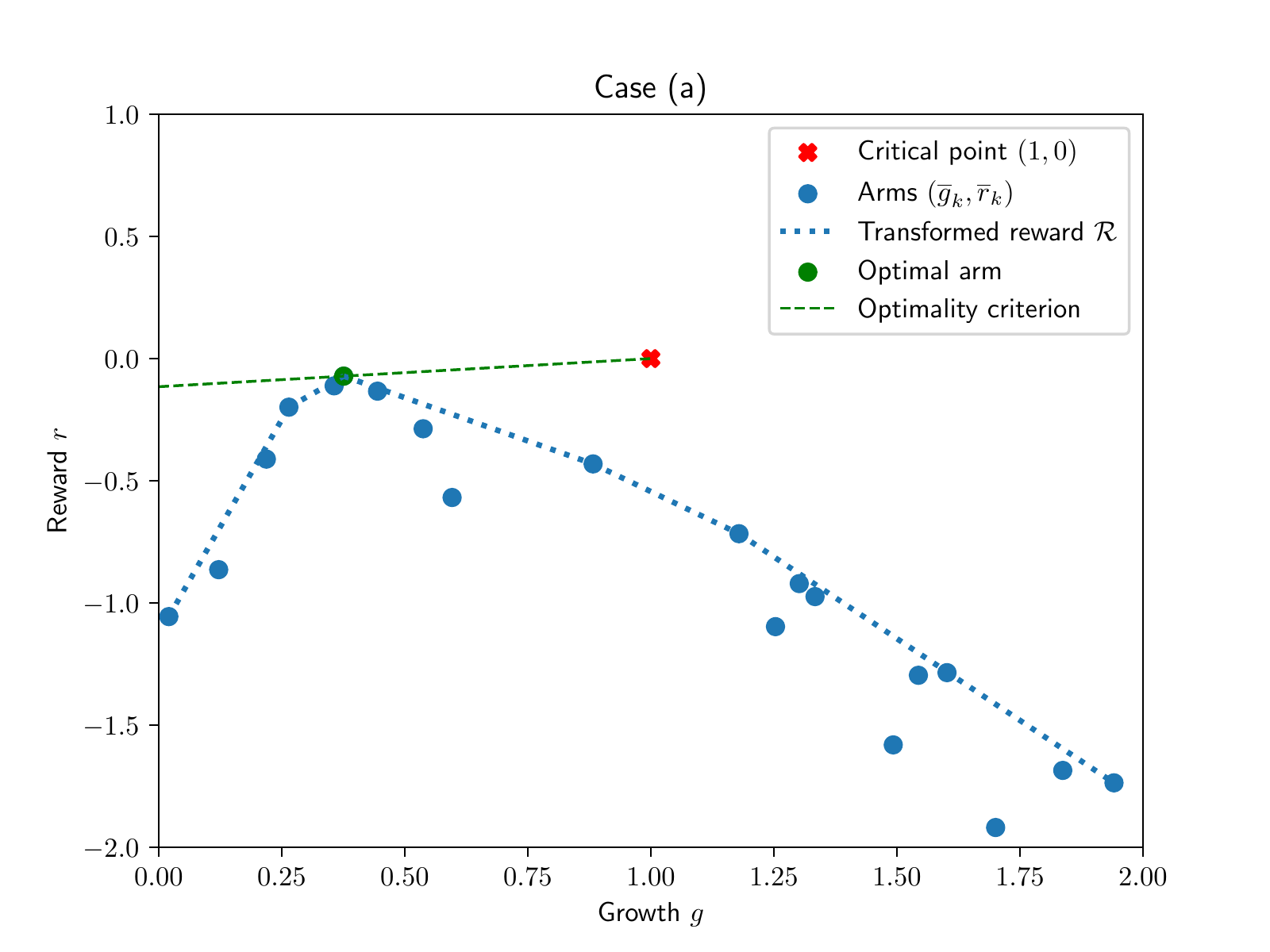}
    \includegraphics[trim = 15pt 5pt 35pt 25pt, clip,width=0.33\textwidth]{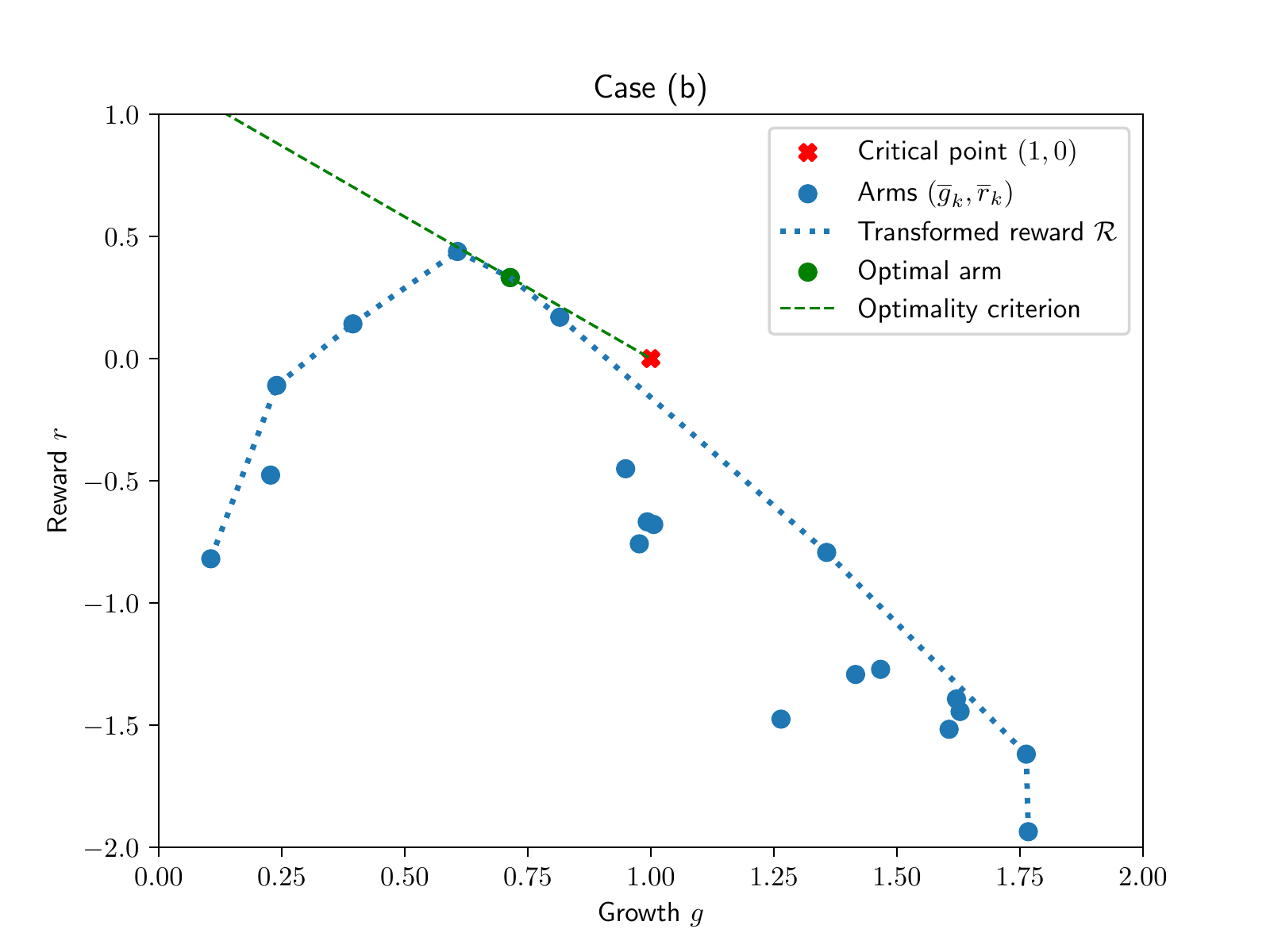}
    \includegraphics[trim = 15pt 5pt 35pt 25pt, clip,width=0.33\textwidth]{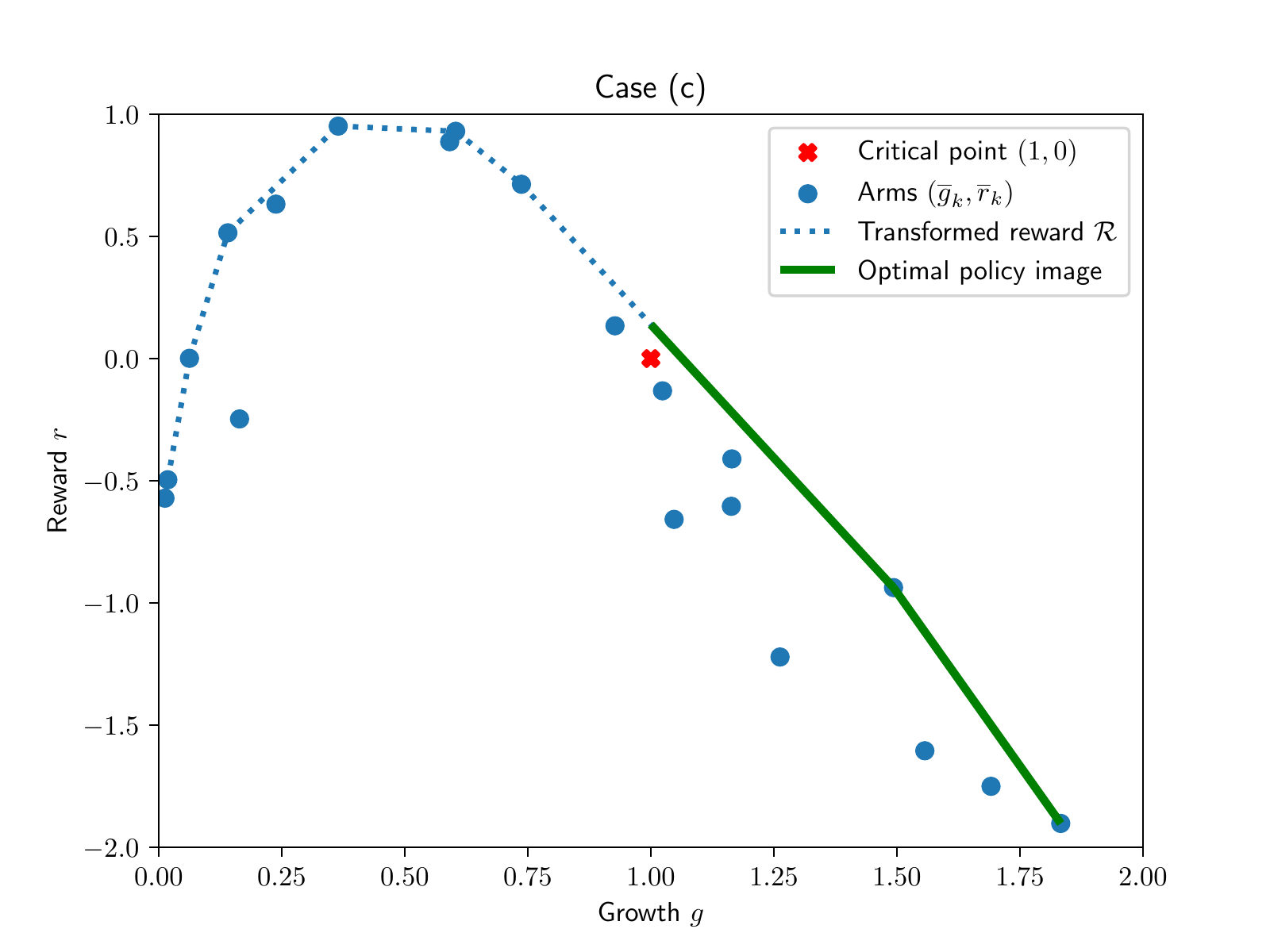}
    \caption{Example of upper convex envelop of the arms parameters ($K=20$).}
    \label{fig:pareto}
\end{figure*}

\section{Analysis}
\label{sec:analysis}
This section analyses the connections between our different MDP definitions \ref{def:PMDP} , \ref{def:OeMDP}, and \ref{def:ROeMDP}. We start with an analysis of the ROeMDP in the form of a series of properties on the transformed reward function, the optimal value function, and the optimal policy. For the sake of space constraint and clarity, most of the proofs have been moved to the supplementary material.

\subsection{ROeMDP properties}
Property \ref{prop:piecewiselinear} states properties of the transformed reward function. Figure \ref{fig:pareto} proposes some visual representations of parameter setting examples with their respective transformed reward function.
\begin{restatable}[Properties of $\mathcal{R}$]{property}{rprop}
    $\mathcal{R}$ is a piece-wise linear concave function. It is the upper convex envelop of the arms parameters $\{(\overline{g}_k,\overline{r}_k)\}_{k\in[K]}$.
    \label{prop:piecewiselinear}
\end{restatable}

Property \ref{prop:optimalityequivalence} states that both modelizations OeMDP and ROeMDP have the same optimal values. As a consequence, we may search for an optimal policy in the simpler ROeMDP and then, retrieve an optimal policy in the OeMDP with the transformed action set.
\begin{restatable}[ROeMDP/OeMDP optimality equivalence]{property}{optimalityequivalence}
    For every optimal policy $\psi^{\mydiese}$ in the OeMDP, there exists an optimal policy $\pi^{\mydiese}$ in the ROeMDP such that $\psi^{\mydiese}(x) \in \Psi(\pi^{\mydiese}(x))$, and we have the optimal values equality: $V_r^{\mydiese}(x) = V_o^{\mydiese}(x)$ for all states $x\in\mathcal{X}_o$.
    \label{prop:optimalityequivalence}
\end{restatable}

Properties \ref{prop:continuity} and \ref{prop:increasing} state remarkable characteristics of the optimal value functions in their MDPs, which are useful to the proofs of the main theorems. More precisely, Property \ref{prop:continuity} proves that the optimal value functions in the ROeMDP are continuous with the respect to the state and the action, and Property \ref{prop:increasing} demonstrates that the optimal value functions in PMDP/OeMDP (and the ROeMDP by consequence) are monotonously increasing or decreasing depending on the MDP parameters.
\begin{restatable}[ROeMDP optimal value function continuity]{property}{continuity}
    In the ROeMDP, the optimal value functions $V_r^{\mydiese}$ and $Q_r^{\mydiese}$ are continuous in $x$ and $g$.
    \label{prop:continuity}
\end{restatable}

\begin{restatable}[OeMDP/PMDP optimal value function monotonicity]{property}{monotonicity}
    When there exists an arm with positive reward (resp. when all arms have a negative reward), the optimal value functions $V^{\mydiese}_o$ or $V^*_p$ are (i) positive (resp. negative), (ii) strictly monotonically increasing w.r.t. $x$ (resp. decreasing), and (iii) concave w.r.t. $x$ (resp. convex). When the highest reward among arms is equal to 0, then $V^{\mydiese}_o=V^*_p=0$.
    \label{prop:increasing}
\end{restatable}

Property \ref{prop:existence} proves that there exists an optimal policy $\pi^{\mydiese}$ such that the ROeMDP actions taken over time are decreasing. However, it does not necessarily mean that $\pi^{\mydiese}$ is a decreasing function of $g$.
\begin{restatable}[Existence of decreasing optimal policy]{property}{existence}
    In the ROeMDP, if $\max_{g\geq 1} \mathcal{R}(g)\leq 0$, or if $\gamma$ is chosen such that $\max_{g\geq \frac{1}{\gamma}} \mathcal{R}(g)>0$, there exists an optimal policy $\pi^{\mydiese}$ that is monotonically decreasing with time: $\forall x, \pi^{\mydiese}(x)\geq \pi^{\mydiese}(\min\{x\pi^{\mydiese}(x),x_{\mytop}\})$.
    \label{prop:existence}
\end{restatable}

From those properties, depending on the parameters of the ROeMDP, we may classify the setting into three different cases:
\renewcommand{\labelenumi}{Case (\alph{enumi})}
\begin{enumerate}[leftmargin=1.5cm]
    \item $\max_{g\in\mathcal{G}} \mathcal{R}(g) \leq 0$ which is equivalent to $\max_{k\in[K]} \overline{r}_k \leq 0$: It means that all rewards are negative and the goal is therefore to diminish the crowd at the least cost. Figure \ref{fig:pareto}(a) illustrates this case.
    \item $\max_{g\in\mathcal{G}} \mathcal{R}(g) > 0$ but $\max_{g\geq 1} \mathcal{R}(g)\leq 0$: it means that it is possible to get a positive return, but impossible to do it in a sustainable way. 
    Figure \ref{fig:pareto}(b) illustrates  this case.
    \item $\max_{g\geq 1} \mathcal{R}(g) > 0$: it means that it is possible to get a positive return in a sustainable way. 
    Figure \ref{fig:pareto}(c) illustrates  this case.
\end{enumerate}

For Cases (a-b), there exists an analytical solution: Theorem \ref{th:below(1,0)} proves that there exists a constant optimal policy $\pi^{\mydiese}$ in the ROeMDP and that there exists $\psi^{\mydiese}\in\Psi(\pi^{\mydiese})$ that is deterministic, \textit{i.e.} selects a single arm with probability 1. 

\begin{restatable}[ROeMDP solution in Cases (a-b)]{theorem}{solutionbelow}
    When $\max_{g\geq 1} \mathcal{R}(g) \leq 0$, no sustainable positive reward is possible, consequently, if $\overline{r}_{\mytop}\geq 0$, or if $\gamma$ is chosen close enough to 1:
    \begin{align}
        \gamma \geq \max_{k\in\mathcal{P}_{< 1}} \frac{\overline{r}_{\mytop} - \overline{r}_{k}}{\overline{g}_{k}\overline{r}_{\mytop} - \overline{r}_{k}},
    \end{align}
    where $\mathcal{P}_{< 1}$ is the set of arms $k$ such that $\overline{g}_k < 1$, then under the optimal policy, the crowd decreases geometrically with time. Furthermore, the optimal policy is to constantly and deterministically play the same arm maximizing the value function:
    \begin{align}
        V^{\mydiese}_r(x) &= x \max_{k \in \mathcal{P}_{<1}} \frac{\overline{r}_k}{1- \gamma\overline{g}_k}.
    \end{align}
    \label{th:below(1,0)}
\end{restatable}

Thus, Cases (a-b) are similar, and the optimal arm may be geometrically interpreted by letting a half-line anchored on the critical point $(\frac{1}{\gamma},0)$ fall on the transformed reward $\mathcal{R}$ curve. The optimal arm is the one that is in contact with the half-line. This is illustrated on Figures \ref{fig:pareto}(a-b) with the dashed green line (here $\gamma$ is set to 1\footnote{In practice, we will always set $\gamma=1$ in Cases (a-b).}).

In Case (c), we set $\gamma$ sufficiently close to 1 to ensure that $\max_{g\geq \frac{1}{\gamma}} \mathcal{R}(g) > 0$. 
Property \ref{prop:existence} proves the existence of a decreasing optimal policy that first ensures the crowd growth, possibly at some cost, until reaching $x_{\mytop}$, where it selects transformed action $g_*=\argmax_{g\geq 1} \mathcal{R}(g)$ ($g_*=1$ on Figure \ref{fig:pareto}(c)).

\subsection{PMDP near-optimality of the ROeMDP optimal policy with model errors}
This theorem states that the ROeMDP formalization allows to find a policy that is near optimal in the true PMDP environment, even with an imperfect model of the ROeMDP environment. Its proof has been kept in the main document because the most technical parts are abstracted into lemmas and corollaries that the interested reader may find in the supplementary material. For clarity, the theoretical results are presented below in order of magnitude. The multiplicative constants may be retrieved by looking at the lemmas, corollaries, and properties the theorem relies on.

\algrenewcommand\algorithmicindent{10pt}
\begin{table*}[t]
\begin{subalgorithm}{.5\textwidth}
\hspace*{1.5em} \textbf{Input:} $x_0$, $x_{\mytop}$, $T$, $\delta$.
\begin{algorithmic}[1]
    \State Initialize history: $\forall k\in[K], h_k=\emptyset$.
    \State Set $\xi = \frac{1}{\sqrt{2}}\max(\dot{g}_{\mytop},\dot{r}_{\mytop}-\dot{r}_{\mybot})\sqrt{\ln\frac{2}{\delta}}$.
    \For{$t=0$ to $T$}
        \State Compute the confidence bounds: 
        \hspace*{\fill} $$\forall k \in[K] \quad \left\{\begin{array}{l}
             CI_k = \xi |h_k|^{-\frac{1}{2}} \\
             g^{\myplus}_k = \frac{1}{|h_k|}\sum_{\tau\in h_k} \dot{g}_\tau + CI_k  \\
             g^{\myminus}_k = \frac{1}{|h_k|}\sum_{\tau\in h_k} \dot{g}_\tau - CI_k  \\
             r^{\myplus}_k = \frac{1}{|h_k|}\sum_{\tau\in h_k} \dot{r}_\tau + CI_k
        \end{array}\right.$$
        \State Compute $\mathcal{R}^{\myplus}$ from $g^{\myplus}_k$ and $r^{\myplus}_k$.
        \If{$\forall g\geq 1$, $\mathcal{R}^{\myplus}(g) \leq 0$}
            \State Set $g^o_k = g^{\sign{r^{\myplus}_k}}_k$.
            \State Select $k_* = \argmax_{k \text{ s.t. } g^o_k < 1}  \frac{r^{\myplus}_k}{1-g^o_k}$.
        \Else
            \State Solve the OeMDP built from $\mathcal{R^{\myplus}}$: $\psi$.
            \State Select: $k_* \sim \psi(x)$.
        \EndIf
        \State Gather a sample: $\tau_i = \langle k_*, \dot{g}, \dot{r}\rangle$
        \State Update: $h_{k_*} \leftarrow h_{k_*} \cup \{\tau_i\}$
    \EndFor
\end{algorithmic}
\caption{Online version}
\label{alg:main}
\end{subalgorithm}
\begin{subalgorithm}{.5\textwidth}
\hspace*{1.5em} \textbf{Input:} $x_0$, $x_{\mytop}$, $T$, $\delta$.
\begin{algorithmic}[1]
    \State Initialize history: $\forall k\in[K], h_k=\emptyset$.
    \For{$t=0$ to $T$}
        \State Initialize selection: $\forall k\in[K], s_k=0$.
        \For{$i=0$ to $x_t$}
            \State Compute the confidence bounds: 
            $$\forall k \in[K] \quad \left\{\begin{array}{l}
                 CI'_k = \xi (|h_k|+s_k)^{-\frac{1}{2}} \\
                 g^{\myplus}_k = \frac{1}{|h_k|}\sum_{\tau\in h_k} \dot{g}_\tau + CI'_k  \\
                 g^{\myminus}_k = \frac{1}{|h_k|}\sum_{\tau\in h_k} \dot{g}_\tau - CI'_k  \\
                 r^{\myplus}_k = \frac{1}{|h_k|}\sum_{\tau\in h_k} \dot{r}_\tau + CI'_k
            \end{array}\right.$$
            \State Compute $\mathcal{R}^{\myplus}$ from $g^{\myplus}_k$ and $r^{\myplus}_k$.
            \If{$\forall g\geq 1$, $\mathcal{R}^{\myplus}(g) \leq 0$}
                \State Set $g^o_k = g^{\sign{r^{\myplus}_k}}_k$.
                \State Select: $k_i = \argmax_{k \text{ s.t. } g^o_k < 1}  \frac{r^{\myplus}_k}{1-g^o_k}$.
            \Else
                \State Solve the OeMDP built from $\mathcal{R^{\myplus}}$: $\psi$.
                \State Select: $k_i \sim \psi(x)$.
            \EndIf
            \State Update: $s_{k_i} \leftarrow s_{k_i} + 1$
        \EndFor
        \State Gather samples: $\forall i\in[x_t], \tau_i = \langle k_i, \dot{g}, \dot{r}\rangle$
        \State Update: $\forall i\in[x_t], h_{k_i}$: $h_{k_i} \leftarrow h_{k_i} \cup \{\tau_i\}$
    \EndFor
\end{algorithmic}
\caption{Batched version}
\label{alg:main-batched}
\end{subalgorithm}
\captionsetup{labelformat=alglabel}
\caption{Upper Confidence Bounds for MAB with crowd externalities}
\end{table*}

\begin{figure*}[t]
    \includegraphics[trim = 5pt 5pt 5pt 5pt, clip, width=0.33\textwidth]{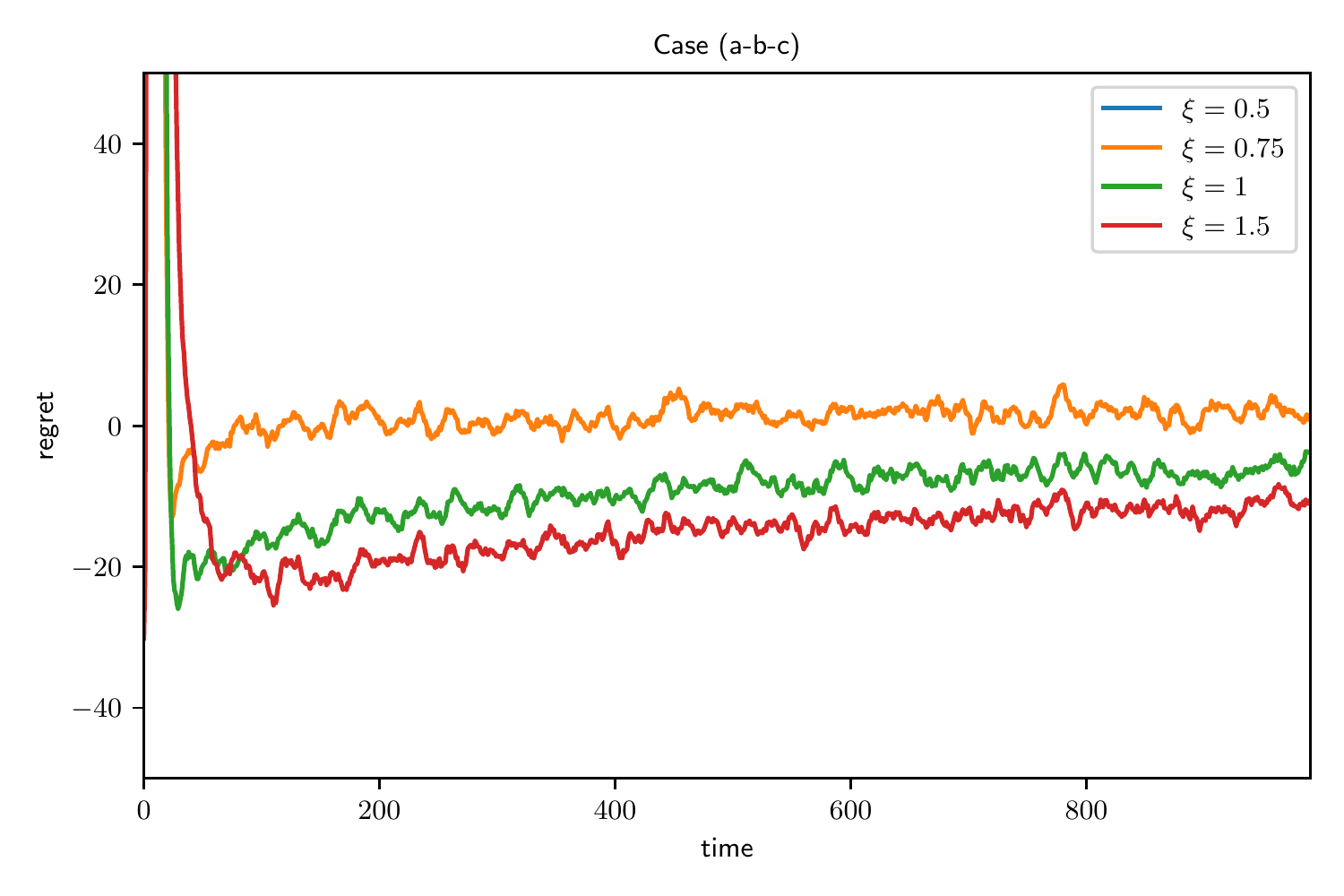}
    \includegraphics[trim = 5pt 5pt 5pt 5pt, clip,width=0.33\textwidth]{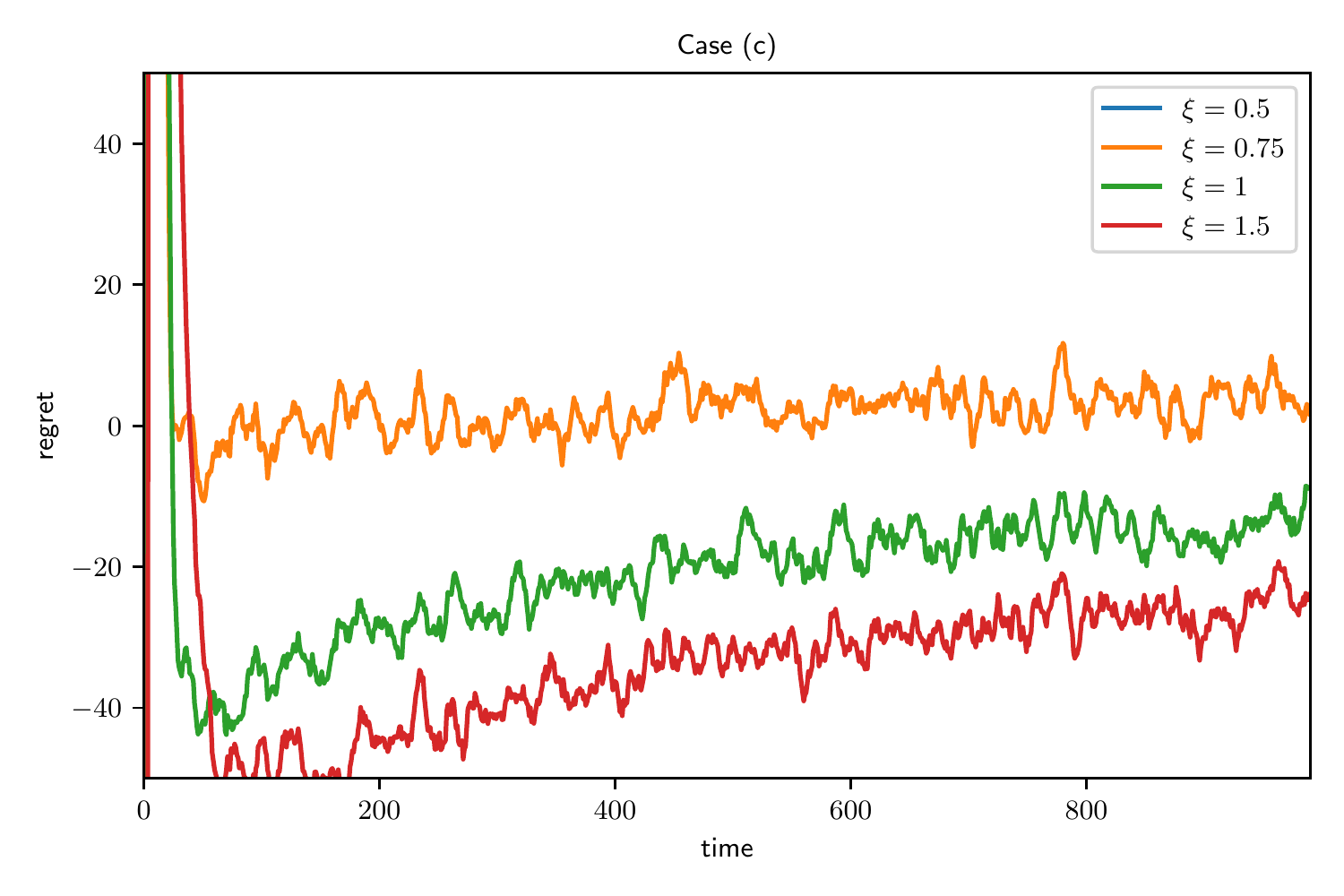}
    \includegraphics[trim = 5pt 5pt 5pt 5pt, clip,width=0.33\textwidth]{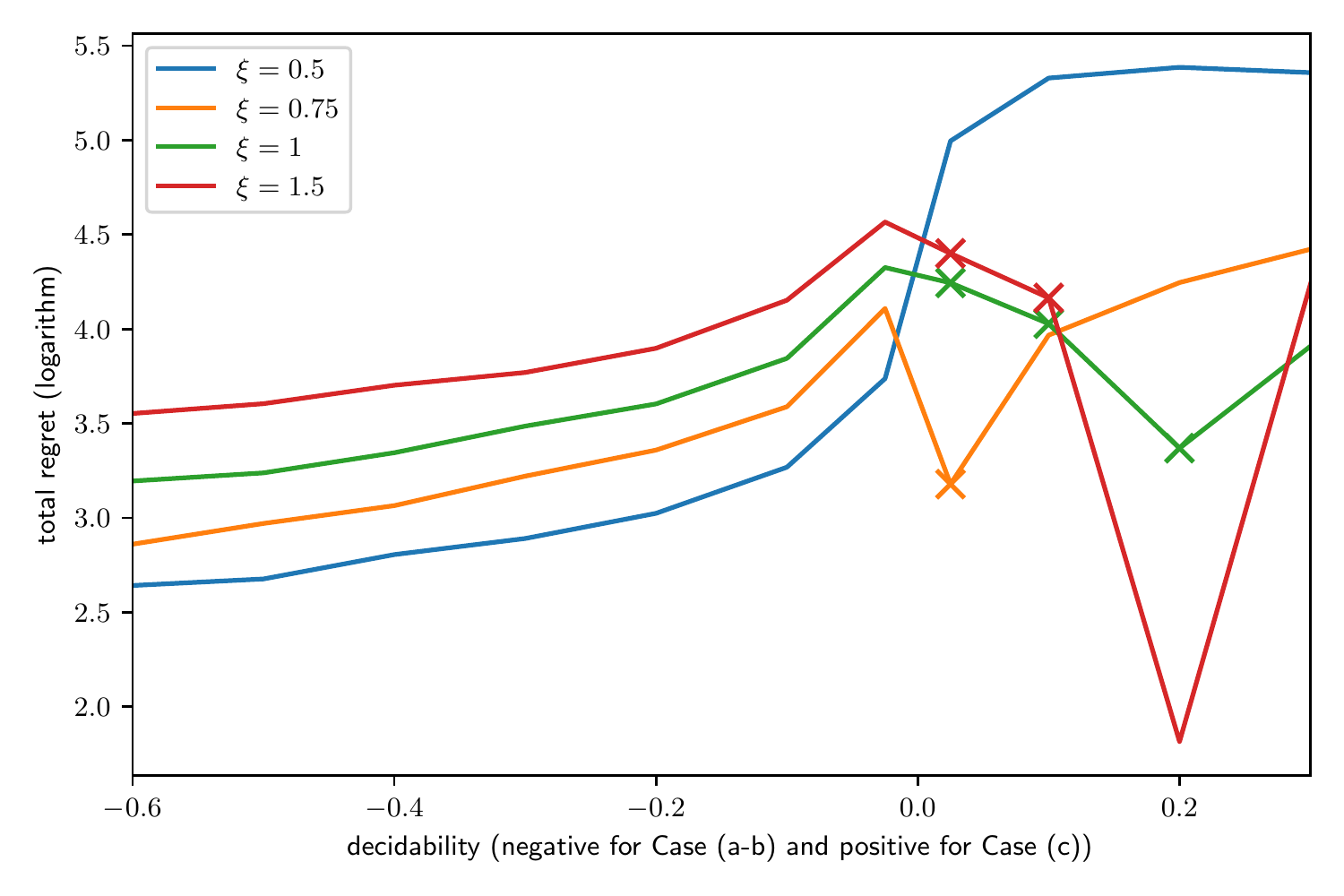}
    \caption{Online experiments. (left) instantaneous regret averaged on all cases, (centre) instantaneous regret averaged on all Case (c), and (right) global regret on a log scale as a function of the decidability (negative for Cases (a-b) and positive for Case (c)), where a cross means that the regret is negative.}
    \label{fig:online}
\end{figure*}

\begin{restatable}[approximate model error on the optimal PMDP value]{theorem}{optimalPMDPerror}
    In the real PMDP environment, the difference between its optimal value and the value of the ROeMDP-optimal policy with estimated parameters obeys the following order of magnitude: 
    \begin{align}
        \mathcal{O}\left(\frac{\dot{g}_{\mytop} V^{\mydiese}_{\mytop}}{1-\gamma}\sqrt{\frac{\ln x}{x}} + \frac{\epsilon V^{\mydiese}_{\mytop}}{1-\gamma}\right),
    \end{align}
    where $x$ is the current crowd, $\dot{g}_{\mytop}$ is the maximal growth, $V^{\mydiese}_{\mytop}$ is the maximal value, and $\epsilon$ is the maximal error of the arms played by the true-PMDP optimal policy and the estimated-ROeMDP optimal policy. \label{th:optimalPMDPerror}
\end{restatable}
\begin{proof}
    The error of control is the difference between the optimal value in the true PMDP environment $V^*_p$ and the value of $\widehat{\psi}$ in the true PMDP environment $V^{\widehat{\psi}}_p$, where $\widehat{\psi} = \Psi(\widehat{\pi})$, and $\widehat{\pi}$ is the policy that is optimal in the ROeMDP built from the imperfect model of the environment. $V^*_p-V^{\widehat{\psi}}_p$ may be broken down into five terms:
    \begin{align}
        \underbrace{V^*_p-V^{\mydiese}_r}_{\textbf{(I)}} + \underbrace{V^{\mydiese}_r - \widehat{V}^{\widehat{\pi}}_r}_{\textbf{(II)}} + \underbrace{\widehat{V}^{\widehat{\pi}}_r - \widehat{V}^{\widehat{\psi}}_o}_{\textbf{(III)}} + \underbrace{\widehat{V}^{\widehat{\psi}}_o - V^{\widehat{\psi}}_o}_{\textbf{(IV)}} + \underbrace{V^{\widehat{\psi}}_o - V^{\widehat{\psi}}_p}_{\textbf{(V)}}.\nonumber
    \end{align}
    
    \textbf{(I)} Corollary \ref{cor:crossoptimal} states that this term is non-positive and may therefore be upper bounded by 0.
    
    \textbf{(II)} This term is the error induced by the misplacement of the upper convex envelop of the arms parameters in the ROeMDP. It is further broken down as follows:
    \begin{align}
        V^{\mydiese}_r - \widehat{V}^{\widehat{\pi}}_r &= \underbrace{V^{\mydiese}_r - V^{\tilde{\pi}}_r}_{\text{(i)}} + \underbrace{V^{\tilde{\pi}}_r - \widehat{V}^{\tilde{\pi}}_r}_{\text{(ii)}} + \underbrace{\widehat{V}^{\tilde{\pi}}_r - \widehat{V}^{\widehat{\pi}}_r}_{\text{(iii)}},
    \end{align}
    where $\tilde{\pi}$ is an optimal policy in the true ROeMDP, under the constraint that $\forall x,$ $\tilde{\pi}(x) \in \mathcal{G}\cap\widehat{\mathcal{G}}$.
    \renewcommand{\labelenumi}{(\roman{enumi})}
    \begin{enumerate}
        \item Lemma \ref{lem:domainpareto} proves that this error is linearly decreasing with the growth domain error $\mathcal{G} - \widehat{\mathcal{G}}$.
        \item Lemma \ref{lem:rewardpareto} proves that this error is linearly decreasing with the upper convex envelop reward error $\big\lVert\mathcal{R}(g) - \widehat{\mathcal{R}}(g)\big\rVert_\infty$.
        \item This term has to be non-positive since $\widehat{\pi}$ is optimal in the estimated ROeMDP.
    \end{enumerate}
    
    As a consequence, the error $V^{\mydiese}_r - \widehat{V}^{\widehat{\pi}}_r$ linearly depends on the model error $\epsilon$ on played arms, either by the true optimal policy or by the target policy.
    
    \textbf{(III)} Property \ref{prop:optimalityequivalence} states that the values are equal, hence, their difference is 0.
    
    \textbf{(IV)} This term is the reduction error. It accounts for the error between the estimated dynamics and the true dynamics in the OeMDP. Lemma \ref{lem:OeMDPmodelerror} proves that the reduction error is bounded as a function of the error on the dynamics estimates over the arms in the image $\widehat{\psi}[\mathcal{X}_o]$ of the trained policy $\widehat{\psi}$. More precisely, $\widehat{V}^{\widehat{\psi}}_o-V^{\widehat{\psi}}_o$ is upper bounded by:
    \begin{align}
         \frac{x\max_{k\in\widehat{\psi}[\mathcal{X}_o]} \lvert\overline{r}_k - \widehat{r}_k\rvert + \gamma V^{\mydiese}_{\mytop}\max_{k\in\widehat{\psi}[\mathcal{X}_o]} \lvert\overline{g}_k - \widehat{g}_k\rvert}{1-\gamma},
    \end{align}
    where $\widehat{g}_k$ (resp. $\widehat{r}_k$) is the expected growth (resp. reward) estimate of Arm $k$.
    
    \textbf{(V)} This term is the OeMDP error: the error made by planning in a deterministic on-expectation environment instead of the real stochastic PMDP environment. Lemma \ref{lem:OeMDPerrorbelow} states that the error decreases exponentially with $x_{\mytop}-x$ in Case (a-b) and Lemma \ref{lem:OeMDPerrorabove} deals with Case (c) to demonstrate an overall upper bound of this error in $\mathcal{O}\left(\frac{\dot{g}_{\mytop} V^{\mydiese}_{\mytop}\sqrt{\ln x}}{(1-\gamma)\sqrt{x}}\right)$, where $x$ is the current crowd, $\dot{g}_{\mytop}$ is the maximal growth, and $V^{\mydiese}_{\mytop}$ is the maximal value in the OeMDP.
\end{proof}

The main result displayed in the abstract and the introduction is retrieved when dependencies in $\gamma$, $V^{\mydiese}_{\mytop}$, and $\dot{g}_{\mytop}$ are omitted. The first term may be interpreted as the amplitude of error due to the PMDP-suboptimality of the on-expectation optimal policy, and the second term as the error due to the model error.


\section{Bandit algorithm}
\label{sec:bandit}

We first propose a fully online (as opposed to batched) algorithm, formally described as Algorithm \ref{alg:main}. The principle consists in considering parameters upper confidence bound and solving this optimistic setting as described in Section \ref{sec:analysis}. So, at every time step, the confidence bounds of the parameters of interest are computed (Line 4). Then, Line 5 computes $\mathcal{R}^{\myplus}$: the transformed reward function for the upper confidence bounds. This allows us to decide whether the problem is known to be of Cases (a-b) with high probability (Lines 6-8). Otherwise, the case of the problem may either be still undetermined or known to be Case (c). Regardless, the OeMDP is solved for $\mathcal{R}^{\myplus}$ and its policy is followed (Lines 10-11). Indeed, even when the case is undetermined, following Case (c) consists in growing the crowd, and therefore speeding up further exploration to determine the case. Thus, similarly to classic UCB~\citep{Auer2002,Auer2010}, all arms are ensured to be played until their optimality is either ruled out or confirmed. However, there is an important difference: once Case (c) is refuted, the crowd geometrically decreases and only little more samples are to be collected. It means that the decision to refute Case (c) is irreversible. As a consequence, horizon $T$ has to be known in advance to select the high probability hyperparameter $\delta$. 

Since our bandit algorithms intend to retrieve the ROeMDP-optimal policy, we are going to use the concept of instantaneous expected regret relative to the policy that is obtained by solving the ROeMDP with the true parameters:
\begin{align}
    \rho_{ins}(\mathfrak{A},t) &= \mathbb{E}_{\pi^{\mydiese}}\left[\sum_{\tau=1}^{x_{t}} \dot{r}_\tau\right] - \mathbb{E}_{\mathfrak{A}}\left[\sum_{\tau=1}^{x_{t}} \dot{r}_\tau\right],
    \label{eq:regret}
\end{align}
where $\pi^{\mydiese}$ is the optimal policy in the ROeMDP with the true parameters $\{\overline{g}_k,\overline{r}_k\}_{k\in[K]}$, $x_t$ is a random variable denoting the size of the population at time $t$, and $\dot{r}_\tau$ is a random variable denoting the reward received from individual $\tau$ at time $t$. The cumulative regret is the discounted sum of instantaneous regret over time. Below, Theorem \ref{th:bandittheorem} provides an upper bound to the asymptotic regret of Algorithm \ref{alg:main}, that guides us, without knowledge on the encountered case, to set $\delta \in \mathcal{O}\left(\frac{1}{T}\right)$, so that the overall expected regret due to parameter estimation would follow an asymptotic regret in $\mathcal{O}(\sqrt{T\ln T})$ in the worst setting (Case (c) with growth of the max-reward arm smaller than 1) and $\mathcal{O}(K\ln T)$ otherwise.

\begin{restatable}[Algorithm \ref{alg:main} expected regret]{theorem}{banditthm}
    The cumulative regret of Algorithm \ref{alg:main} is upper bounded by an error term that decays with the following order of magnitude as a function of the number of arms $K$, the horizon $T$, and the high probability hyper-parameter $\delta$:
    \begin{enumerate}[leftmargin=2cm]
        \item[Case (a-b)] $\mathcal{O}\left(K\ln\frac{1}{\delta} + K\delta T\right)$,
        \item[Case (c)] $\left\{\begin{array}{l}
            \mathcal{O}\left(K\ln\frac{1}{\delta} + K\delta T\right)\text{ if } g_{k_*}>1, \\
            \quad\quad \text{with }k_* = \argmax_{k\in[K]} r_k \\
            \mathcal{O}\left(K\ln\frac{1}{\delta} + K\delta T + \sqrt{T\ln\frac{1}{\delta}}\right),\\
            \quad\quad\text{ otherwise.}
        \end{array} \right.$
    \end{enumerate}
    \label{th:bandittheorem}
\end{restatable}


The batched version is formalized in Algorithm \ref{alg:main-batched}. The conversion is simple: whereas the outcomes of the arm selections are observed by batches, the selection of arms itself is known and the corresponding confidence interval may be updated. Another difference has to be noted: the number of samples is dependent of the size of the batch and not only the horizon $T$. Still, the nature of the Case (c) policy is such that the maximum crowd is quickly reached (the crowd grows geometrically), point from which the maximal number of samples remaining to be collected until the horizon is easy to upper bound: $(T-t)x_{\mytop}$.

\begin{figure*}[t]
    \includegraphics[trim = 5pt 5pt 5pt 5pt, clip, width=0.33\textwidth]{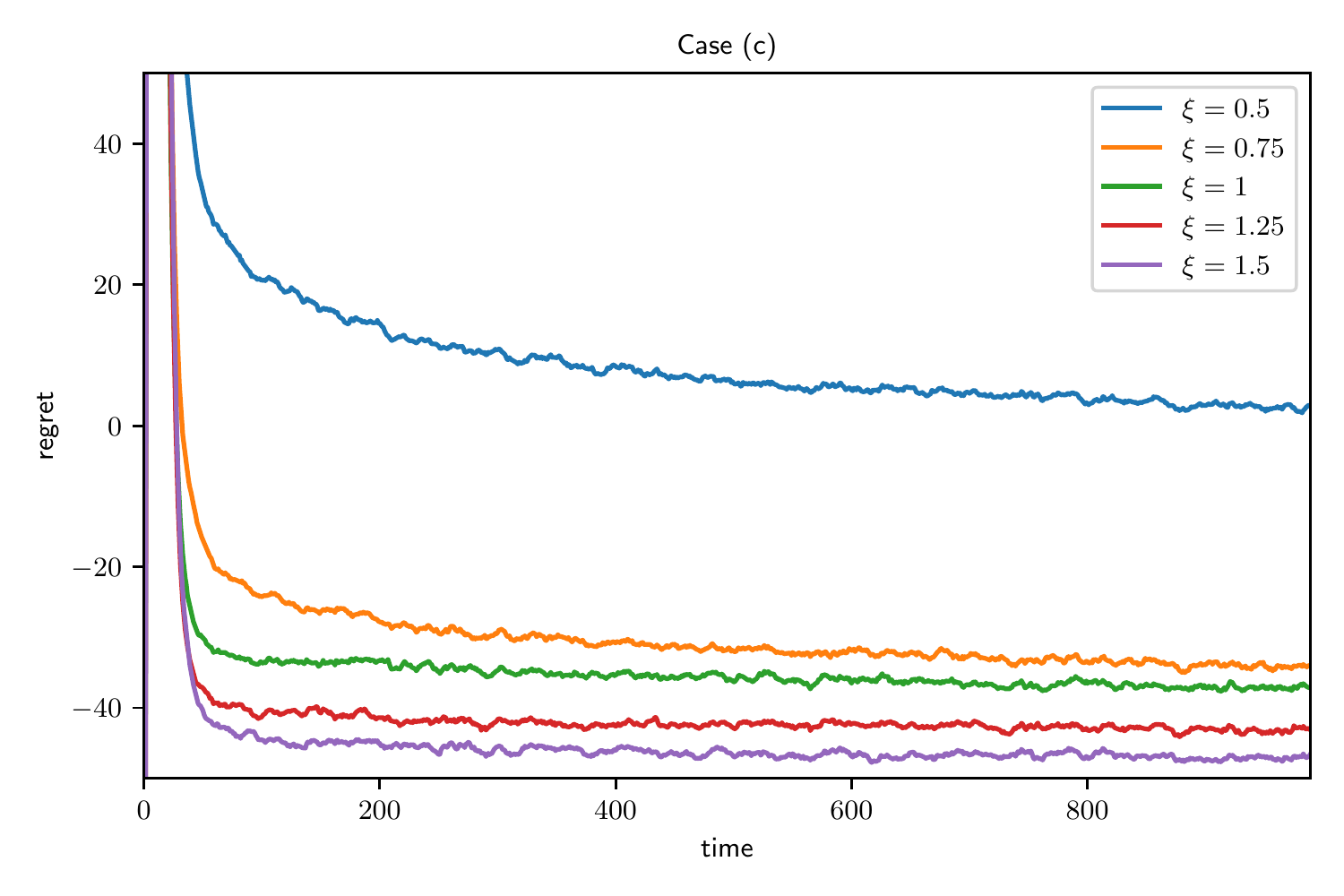}
    \includegraphics[trim = 5pt 5pt 5pt 5pt, clip,width=0.33\textwidth]{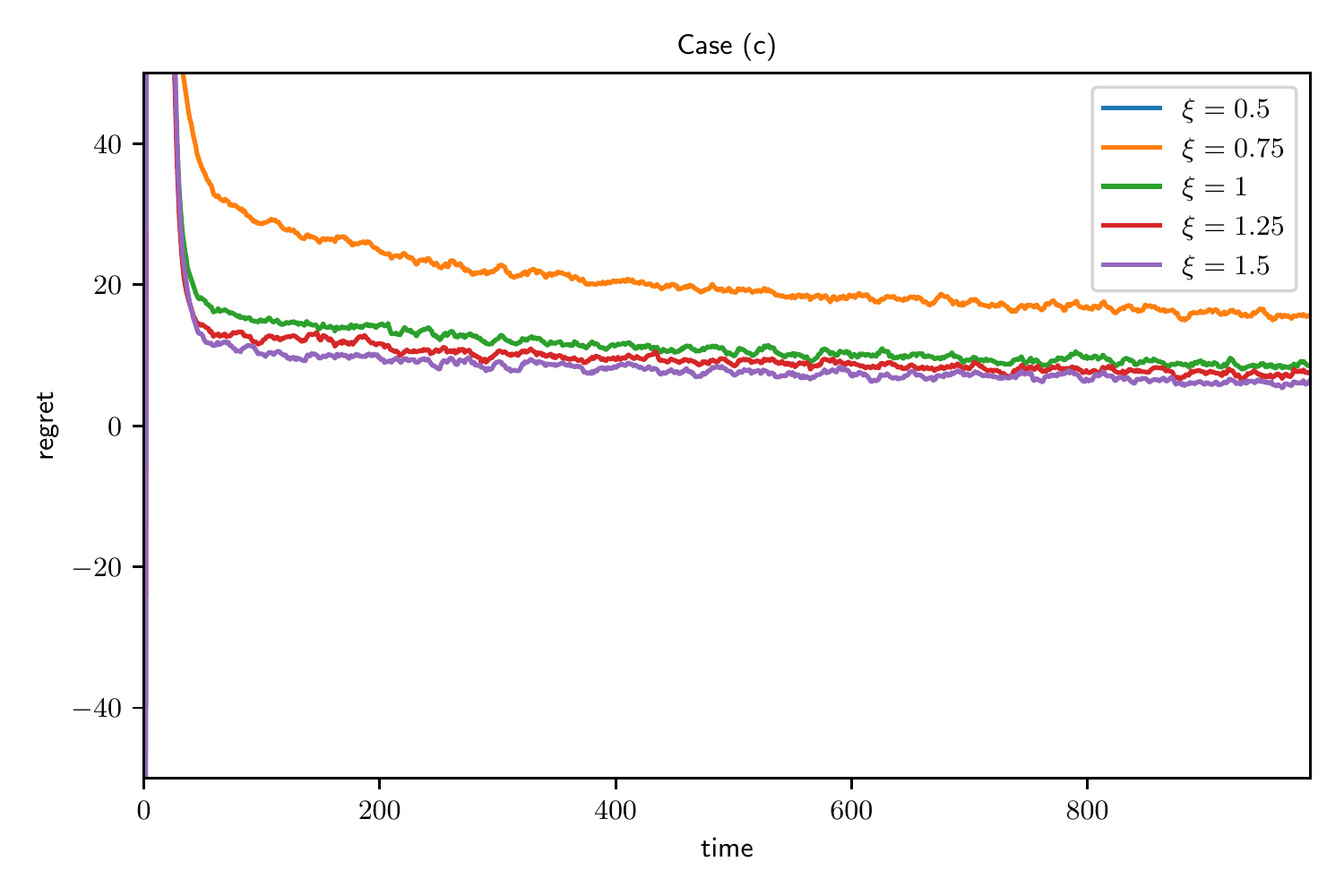}
    \includegraphics[trim = 5pt 5pt 5pt 5pt, clip,width=0.33\textwidth]{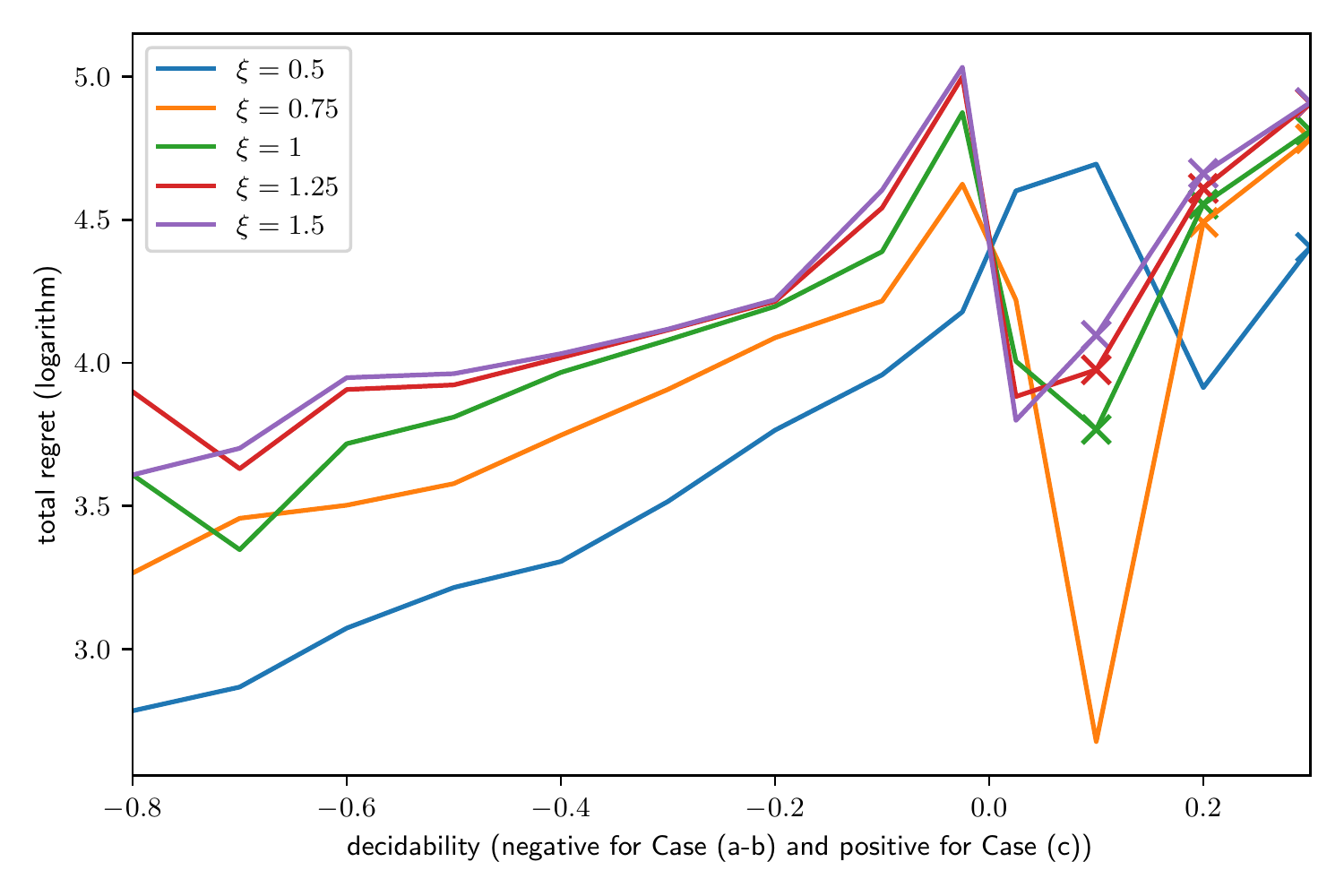}
    \caption{Batched experiments. (left) instantaneous regret averaged on all Case (c), (centre) instantaneous regret averaged on Case (c) when no run lost all crowd, and (right) cumulative regret on a log scale as a function of the decidability (negative for Cases (a-b) and positive for Case (c)), where a cross means that the cumulative regret is negative.}
    \label{fig:batched}
\end{figure*}

\section{Numerical analysis}
\label{sec:experiments}
In order to validate our theoretical findings, we use a generator of problems described in Algorithm \ref{alg:problemgenerator} in Supplementary Material with maximal crowd $x_{\mytop}=10,000$ and horizon $T=1,000$. This generator has been tuned in order to generate an interesting distribution of problems, \textit{i.e.}:
\begin{itemize}
    \item that have approximately 10\% chance to be of Case (a), 40\% of Case (b), and 50\% of Case (c),
    \item that have transformed reward functions composed of several segments, 
    \item and that have a maximal reward that is obtained for growth below 1. 
\end{itemize}

Figures \ref{fig:pareto}(a-c) show settings created with this generator. During the design of the task generator, it quickly appeared that some are easier than others. In order to account for this and to analyze to performance correlation with the difficulty, we define hereafter the decidability: a generalization of the notion of gap classically used in MAB.

\begin{definition}[Decidability]
    Given a task characterized by its arms parameters, we define the decidability as the distance between its transformed reward function and point $(1,0)$.
\end{definition}

Indeed, the further it is from this point, the easier it is to determine whether the problem is of Cases (a-b) or Case (c). 

The benchmark of our experiments only includes our UCB-based algorithm because, to the best of our knowledge, no algorithm in the literature is able to take into account the dual growth-reward feedback. As a consequence, classic MAB/BMAB would eventually select the arm with the highest reward, which is unlikely to have a growth higher than 1, and therefore would deplete the crowd. RL algorithm, as well, are not designed to deal with settings such as ours, where only one trajectory is allowed and some states are final. Consequently, RL algorithms would explore states with low crowds and risk crowd depletion. Finally, we considered a Thompson Sampling (TS) formulation of the ROeMDP parameter exploration/exploitation~\citep{Chapelle2011,Kaufmann2012}. However, the results were so poor that we opted to not report them and to provide instead the following qualitative explanation on why it cannot work as well as UCB in our setting. UCB is by nature an optimistic algorithm. Its optimism may sometimes be detrimental, but is actually virtuous in our setting because it naturally urges the algorithm to pull arms with high growths, which mitigates the risk of depleting the crowd. In contrast, TS indifferently accounts for uncertainty, sometimes optimistically, but also sometimes pessimistically, which implies in our setting to commit to Case (a-b) too early, and therefore to deplete the crowd. We do not claim that there does not exist any efficient TS-based algorithm for our setting but the adaptation is not as straightforward as for UCB, at the very least.

Figure \ref{fig:online} shows the results for the online experiments. On the left, we have the averaged instantaneous regret as a function of time for all cases and 4 values of $\xi$ ($\xi=0.5$ is off chart). Indeed, it is more practical to directly perform a hyper-parameter search on $\xi$ (defined on Line 2 of Algorithm \ref{alg:main}), rather than $\delta$. It is interesting to observe that the instantaneous regret as defined in Equation \ref{eq:regret} gets negative: the bandit algorithm does better in practice than the optimal point it is looking for. In the middle, we isolate Case (c), which reveals to be responsible for such an unexpected result. This is explained as follows: the bandit uses optimistic values for the unknown parameters of the problem and chooses the best trade-off on it such that $g=1$ when the population is already maximal. It means that in practice, it aims at growth that are a little smaller than required, meaning that it hits the $x_{\mytop}$ ceiling less frequently and remains more likely at a safe distance from it. In contrast the true optimal will hit $x_{\mytop}$ more often and lose some expected benefit from it. On the right, we observe the $\log_{10}$ averaged total regret per decidability level. We observe that smaller $\xi$ benefits to Cases (a-b) while this is the opposite for Case (c). This was expected since a higher $\xi$ means that Case (c) strategy has to be followed for a longer time.

Figure \ref{fig:batched} shows the results for the batched experiments. Once again, we once again notice on the left figure that the instantaneous regret gets negative. However, contrarily to the online setting, the instantaneous regret does not tend to 0 asymptotically. This negative regret is explained by the fact that, in some settings, the optimal growth is very small and incurs a risk of crowd depletion. In contrast, the bandit algorithm is reward optimistic, which makes it select much higher growth targets. The middle figure represents the results when none of the runs suffered a crowd depletion, where the instantaneous regret is positive, and tends to 0 asymptotically. The right figure displays the total reward as a function of the decidability, with similar results as in the online experiments.

\section{Conclusion}
\label{sec:conclusion}
We tackled the problem of Batched Multi-Armed Bandits in an environment where the future affluence depends on the past arm selections. We built an approximate formulation of the problem in order to make it tractable. We proved the near-optimality of the approximate solution, and expressed its sensitivity to errors in the parameters. We identified three cases: (a) all arm rewards are negative, (b) it is impossible to maintain the crowd and get positive expected rewards, and (c) it is possible to get positive expected rewards while maintaining the crowd. We designed a novel UCB algorithm that allows to grow the crowd as long as the case is undefined and show that this algorithm suffers a regret in $\mathcal{O}(\max(K\ln T,\sqrt{T\ln T}))$ as compared to the approximate solution. We ran experiments that reveal that the bandit often performs better than the approximate solution in the . This unexpected phenomenon is explained by the fact that UCB's optimism has a positive effect on the setting.

\paragraph{Perspectives:} We studied asymptotic regret in the general case, but focusing on the regret in Cases (a-b) may have a greater impact. For instance, management of public health strategies facing epidemy could be cast into our model~\citep{Libin2018}, but our analysis and algorithms would be inefficient at dealing with such settings where the case is known. More generally, we believe that in practice, information about arms is often known: for instance, Arm $k$ is known to yield more reward than Arm $k'$, but generates less growth, and it would be more practical to be able to design algorithm that could take advantage of such prior knowledge. Finally, we have empirical evidence that the ROeMDP solution could be improved by staying at safe distance from $x_{\mytop}$. Formal analysis would be welcome.

\newpage
\bibliographystyle{apalike}
\bibliography{biblio}

\newpage
\onecolumn
\appendix
	
\newenvironment{changemargin}[2]{%
\begin{list}{}{%
\setlength{\topsep}{0pt}%
\setlength{\leftmargin}{#1}%
\setlength{\rightmargin}{#2}%
\setlength{\listparindent}{\parindent}%
\setlength{\itemindent}{\parindent}%
\setlength{\linenumbersep}{10pt-#1}
\setlength{\parsep}{\parskip}%
}%
\item[]}{\end{list}}


\section{Broader Impact}
\label{sec:broaderimpact}
Our work focuses on discovering optimal mitigation between immediate rewards/costs and future spread/containment of the popularity of a system. Our initial motivation for this work is centered around service delivery popularization in a sustainable way, \textit{i.e.} while making it profitable. We endeavour below to enumerate the potential positive (+) and negative (--) societal impacts:
\begin{itemize}
    \item[(+)] \textbf{Stronger economy:} this was our primary objective and we prove that our algorithm makes sure that services without beneficial margins are no longer sustained.
    \item[(+)] \textbf{Faster response to pests:} further work in the same direction but a stronger focus on the Case (a) could be applied to public response to epidemics, locust, or organic pollution (\textit{e.g.} spreading algae).
    \item[(--)] \textbf{Unfairness:} our algorithm chooses arms regardless individual identity, and decides global policies for the best of all, which may, and almost certainly will induce discrimination: \textit{e.g.} Arm 1 would be preferred to Arm 2 because it offers a better service to a majority, even though it is worse for a minority \citep{Thomas2019}.
    \item[(--)] \textbf{Poorly designed reward/cost:} our algorithm optimizes the behaviour in order to maximize a reward function. In practice, the design of the reward function is often an inextricable task: how to mix heterogeneous objectives such as monetary expenditure/income, human casualties (diseased, wounded, deaths), environmental debts/benefits, etc. Some of these effects may only be measured years after, and often partially~\citep{Orseau2016}. 
    \item[(--)] \textbf{Ill-intentioned objectives:} our algorithm could be used for ideological purposes: \textit{e.g.} optimize the spreading of fake news or corrupted ideas, to assist an agenda. Like any tool, it may be used for wrong purposes.
\end{itemize}

\section{Generator of problems}
\begin{figure}[h]
\hspace*{1.5em} \textbf{Input:} $K$.
\begin{algorithmic}[1]
    \State $\alpha \sim \mathcal{U}(0,1).$
    \State $\forall k, \overline{g}_k \sim \mathcal{U}(0,2).$
    \State $\forall k, g_k = \mathcal{G}\left(\frac{\overline{g}_k}{\overline{g}_k+1}\right).$
    \State $\forall k, \overline{r}_k \sim \left.\begin{array}{c}(0.6+0.7\alpha)(\mathcal{U}(0,1)-|2\overline{g}_k-1|)\\-0.5+1.47\alpha \end{array} \right.$
    \State $\forall k, r_k = 4\mathcal{B}(\frac{\overline{r}_k+2}{4}) - 2$
\end{algorithmic}
\captionsetup{labelformat=alglabel}
\caption{Generator of problems}
\label{alg:problemgenerator}
with $\mathcal{U}$, $\mathcal{G}$, and $\mathcal{B}$, respectively denote the uniform, geometric and Bernoulli distributions.

\end{figure}

\newpage
\section{Proofs}

\rprop*
\begin{proof}
    $\mathcal{R}(g)$ may be interpreted as the upper convex envelop of the arms parameters: the expected growth versus the expected reward. Since, there are a finite number of arms, the upper convex envelop must be piece-wise linear and concave.
\end{proof}

\optimalityequivalence*
\begin{proof}
    $\mathcal{R}$ is the upper convex envelop of the points $(\overline{g}_k, \overline{r}_k)$ formed by the $K$ arms. The transition function being entirely determined by the choice of $g$, all optimal action $a$ must belong to $\Psi(g)$, meaning that any optimal policy $\psi^{\mydiese}$ of the OeMDP implements a policy $\pi$ in the ROeMDP. We may infer that $V_o^{\mydiese} = V_r^{\pi} \leq V_r^{\mydiese}$.
    
    Conversely, any optimal policy $\pi^{\mydiese}$ in the ROeMDP may be implemented in the OeMDP by some policy $\psi$ that is a mixture of the two arms surrounding it on the upper convex envelop of the arms parameters. We may infer that $V_r^{\mydiese} = V_o^{\psi} \leq V_o^{\mydiese}$.
    
    From both inequalities, we may conclude that $V_r^{\mydiese} = V_o^{\mydiese}$.
\end{proof}

\continuity*
\begin{proof}
    We first prove the continuity of the optimal value functions $V$ and $Q$ with respect to $x$:
    \begin{align}
        \lim_{\epsilon \rightarrow 0} \left[V(x) - V(x-\epsilon)\right] &\leq  \lim_{\epsilon \rightarrow 0} \left[V(x) - Q(x-\epsilon,\pi(x))\right]\label{eq:optimality1} \\
        &=  \lim_{\epsilon \rightarrow 0} \left[x\mathcal{R}(\pi(x)) - (x-\epsilon)\mathcal{R}(\pi(x)) + \gamma V(\min(x\pi(x),x_{\mytop}))- \gamma V(\min((x-\epsilon)\pi(x),x_{\mytop})\right],
    \end{align}
    where inequality \ref{eq:optimality1} is obtained because $V$ is optimal. We further upper bound and expand $V(x\pi(x))- V((x-\epsilon)\pi(x))$ iteratively and obtain:
    \begin{align}
        \lim_{\epsilon \rightarrow 0} \left[V(x) - V(x-\epsilon)\right] &\leq  \lim_{\epsilon \rightarrow 0} \epsilon\left[\sum_{t=0}^\infty \gamma^t\mathcal{R}(\pi(x^{(t)}))\prod_{t'=0}^{t-1}\pi(x^{(t')})\right] \\
        \text{with} \: x^{(t+1)}=\min\left(x^{(t)}\pi(x^{(t)}),x_{\mytop}\right), &\quad \text{and }\: x^{(0)}= x
    \end{align}
    
    We now prove that the term inside the brackets is finite:
    \begin{align}
        \mathcal{R}(\pi(x^{(t)}))&\leq \max_{k\in[K]} \overline{r}_k \\
        \prod_{t'=0}^{t-1}\pi(x^{(t')})&\leq \frac{x_{\mytop}}{x} 
    \end{align}
    
    We conclude with an upper bound of the limit:
    \begin{align}
        \lim_{\epsilon \rightarrow 0} \left[V(x) - V(x-\epsilon)\right] &\leq  0.
    \end{align}
    
    We can similarly prove the mirrored inequality: 
    \begin{align}
        \lim_{\epsilon \rightarrow 0} \left[V(x-\epsilon)- V(x)\right] &\leq  0,
    \end{align}
    and conclude the proof of the continuity of the optimal value function with respect to $x$.
    
    The continuity of the action-state value function with respect to $x$ and $a$ follows directly from the following expansion, only containing functions that are continuous in $x$ and $a$:
    \begin{align}
        Q(x,a) &=  x \mathcal{R}(a) + \gamma V(\min(xa,x_{\mytop}))
    \end{align}
\end{proof}
    
\monotonicity*
\begin{proof}
    When the highest reward among arms is equal to 0, it is impossible by design to yield positive rewards. The values of always selecting a 0-reward arm are trivially 0, which is therefore optimal.

    \textbf{(i)} The optimal policy has a value that is larger than the one always selecting the positive arm, which must have a positive value. Conversely, if there is no positive reward, then the value cannot be positive.
    
    \textbf{(ii)} For brevity, we use the same notations for $V^{\mydiese}_o$ and $V^*_p$: $V$. When there exists a positive arm, the optimal value in $x+\delta x$ has to be larger than the value of the policy copying $\pi(x)$ for the a subcrowd of size $x$, and takes the maximum immediate reward for the remaining of the crowd $\delta x$:
    \begin{align}
        V(x+\delta x) &\geq V(x) + \delta x \max_{k\in[K]} \overline{r}_k.
    \end{align}
    $\max_{k\in[K]} \overline{r}_k$ has to be positive since we assume that there exists an arm with a positive expected reward. 
    
    Conversely, when there is no positive arm, we have:
    \begin{align}
        V(x+\delta x) &\leq V(x) + \delta x \max_{k\in[K]}\overline{r}_k + \gamma \max_{x'} V(x').
    \end{align}
    We know that the second term is negative and the last term is non-positive, therefore the strict decreasing property is proven.
    
    \textbf{(iii)} When there exists a positive arm, let us consider the decomposition of the value into $n$ units as follows:
    \begin{align}
        V(x) &= \sum_{i=1}^{n} \left\{V\left(\frac{i}{n}x\right)-V\left(\frac{i-1}{n}x\right)\right\} + \underbrace{V(0)}_{=0}\\
        &=  \sum_{i=1}^{n} U_i \quad\text{with}\quad U_i =V\left(\frac{i}{n}x\right)-V\left(\frac{i-1}{n}x\right)\\
    \end{align}
    $(U_i)$ has to be a decreasing sequence, otherwise, a reordering $\sigma$ of the sequence $(U_{\sigma(i)})$ would yield higher values, which is inconsistent with the optimality assumption made on the values. As a consequence, for any $\delta x$, we have:
    \begin{align}
        V(x+\delta x) - V(x) \leq V(x) - V(x-\delta x) \\
        \Leftrightarrow \quad\quad V(x) \geq \frac{1}{2}\left(V(x+\delta x)+V(x-\delta x)\right),
    \end{align}
    which is sufficient to conclude that $V(x)$ is concave.
    
    Conversely, when there is no positive arm, the same proof may be developed, but this time the constraint in size has a positive effect on the value: it prevents the crowd from growing more than what it should and therefore saves subsequent negative reward. As a consequence, the value functions are convex in this case.
\end{proof}

\existence*
\begin{proof}
    When $\max_{g\geq 1} \mathcal{R}(g)\leq 0$, the problem is trivially solved by Theorem \ref{th:below(1,0)}. The policy is constant, therefore decreasing in its broad sense.
    
    In the complementary case, $\max_{g\geq 1} \mathcal{R}(g) > 0$, we may choose $\gamma$ such that: $\max_{g\geq \frac{1}{\gamma}} \mathcal{R}(g) > 0$. Now, we look at the situation in 
    \begin{align}
        g_*= \argmax_{g'\geq \frac{1}{\gamma}} \mathcal{R}(g').
    \end{align}
    
    In a first step, we prove that the optimal policy satisfies $\pi^{\mydiese}(x)\geq g^*$ for all $x$ \textbf{(I)}. In a second step, we prove that, if $x''\geq xg_*^2$, then the optimal way to reach $x''$ after two action $g_1$ and $g_2$ implies that $g_1\geq g_2$ \textbf{(II)}.
    
    
    \textbf{(I)} We prove here that for all $x$, the optimal policy $\pi^{\mydiese}(x)$ is necessarily larger than $g^*$. To do so, we assume $g<g_*$ and prove $Q^{\mydiese}_r\left(x,g\right) \leq Q^{\mydiese}_r\left(x,g_*\right)$. We write that $\mathcal{R}(g')=\alpha g' + \beta$ in the lower vicinity of $g_*$. Since $g_*$ is maximal, $\alpha \geq 0$:
    \begin{align}
        Q^{\mydiese}_r\left(x,g\right) - Q^{\mydiese}_r\left(x,g_*\right) &= x\left(\mathcal{R}(g)-\mathcal{R}\left(g_*\right)\right) + \gamma \left(V^{\mydiese}_r\left(xg\right) - V^{\mydiese}_r\left(xg_*\right)\right) \\
        &\leq \underbrace{x\alpha\left(g-g_*\right)}_{\text{by concavity }\mathcal{R}(g) \leq \alpha g + \beta} + \underbrace{\gamma \left(\frac{g}{g_*}V^{\mydiese}_r\left(xg_*\right) - V^{\mydiese}_r\left(xg_*\right)\right)}_{V^{\mydiese}_r\left(xg\right)\leq\frac{g}{g_*} V^{\mydiese}_r\left(xg_*\right)\text{ by Corollary \ref{cor:lipschitzness}}} \\
        &= x\alpha\left(g-g_*\right) + \frac{\gamma}{g_*} \left(g-g_*\right)V^{\mydiese}_r\left(xg_*\right)\\
        &= \underbrace{\left(g-g_*\right)}_{<0\text{ by assumption}}\left(\underbrace{x\alpha}_{\geq 0} + \frac{\gamma}{g_*}\underbrace{V^{\mydiese}_r\left(xg_*\right)}_{\geq 0}\right).
    \end{align}
    This concludes the proof that the optimal value of $g_*$ is always greater than the optimal value of $g<g_*$, and therefore that $\pi^{\mydiese}(x)\geq g_*$ for all $x$.
    
    \textbf{(II)} We consider two states $x$ and $x''\geq xg_*^2$ and the optimal way to go from $x$ to $x''$ in two steps. In particular, $x''$ could be chosen to be equal to $x\pi^{\mydiese}(x)\pi^{\mydiese}(x\pi^{\mydiese}(x))$, and infer properties over $\pi^{\mydiese}$ but this study is not limited to it. We will use the notation of the double-action value function $Q^{\mydiese}_r(x,g_1,g_2)$ which is the optimal value of taking action $g_1$ followed with action $g_2$ from state $x$. Its Bellman equation is:
    \begin{align}
        Q^{\mydiese}_r(x,g_1,g_2) &= x\mathcal{R}(g_1) + \gamma xg_1\mathcal{R}(g_2) + \gamma^2 V^{\mydiese}_r(xg_1 g_2) \\
        Q^{\mydiese}_r\left(x,g,\frac{x''}{gx}\right) &= x\mathcal{R}(g) + \gamma xg\mathcal{R}\left(\frac{x''}{gx}\right) + \gamma^2 V(x'') \\
        \frac{\partial Q^{\mydiese}_r}{\partial g}\left(x,g,\frac{x''}{gx}\right) &= x\frac{\partial \mathcal{R}}{\partial g}(g) + \gamma x\mathcal{R}\left(\frac{x''}{gx}\right) + \gamma xg\frac{\partial \mathcal{R}}{\partial g}\left(\frac{x''}{gx}\right) \\
        &= x\alpha + \gamma x \left(\alpha'\frac{x''}{gx} + \beta' + g\alpha'\frac{\partial \left(\frac{x''}{gx}\right)}{\partial g}\left(g\right) \right)\\
        &= x\alpha + \gamma x \left(\alpha'\frac{x''}{gx} + \beta' - g\alpha'\frac{x''}{g^2x} \right)\\
        &= x(\alpha + \gamma \beta'),\label{eq:alphabeta'}
    \end{align}
    where $\mathcal{R}(g) =\alpha g + \beta$ in the vicinity of $g$, and $\mathcal{R}(g) =\alpha' g + \beta'$ in the vicinity of $\frac{x''}{gx}$. Because we know that the upper convex envelop is concave in addition of being piece-wise linear, we know that:
    \begin{align}
        \sign\left(g-\frac{x''}{gx}\right) =  -\sign\left(\alpha-\alpha'\right) = \sign\left(\beta-\beta'\right),
    \end{align}
    and we can conclude that the partial derivative of $Q^{\mydiese}_r(x,g,\frac{x''}{gx})$ with respect to $g$ is decreasing and piece-wise constant at each change of segment in the upper convex envelop. It means that that $Q^{\mydiese}_r(x,\frac{x''}{gx})$ is continuous, piece-wise linear and concave as a function of $g$.
    
    Since $g_*$ is maximal, and since $\mathcal{R}$ is concave, $\mathcal{R}$ must be decreasing in $\sqrt{\frac{x''}{x}}\geq g_*$ (by assumption). If we write $\mathcal{R}(g) =\alpha g + \beta$ in the vicinity of $\sqrt{\frac{x''}{x}}$, then we observe that $\alpha$ is non-positive, and again by concavity, that $\alpha g_* + \beta$ must be positive. As a consequence of both observation, we can infer that $\frac{\alpha}{\gamma} + \beta$ must be positive, and therefore that $\alpha + \gamma \beta > 0$, which in turn implies that the partial derivative is positive in $\sqrt{\frac{x''}{x}}$. Thus, we proved that, if $\sqrt{\frac{x''}{x}} \geq g_*$, $Q^{\mydiese}_r(x,g,\frac{x''}{gx})$ reaches its maximum for a value higher than $\sqrt{\frac{x''}{x}}$. 
\end{proof}
    
\solutionbelow*
\begin{proof}
    We compute the value $V_a$ of constantly repeating action $a < 1$:
    \begin{align}
        V_a(x) &= x\mathcal{R}(a) + \gamma V_a(xa) \\
        &= x\sum_{t=0}^\infty (a\gamma)^t\mathcal{R}(a)  \\
        &= x\frac{\mathcal{R}(a)}{1-a\gamma}
    \end{align}
    
    Since $\mathcal{R}(a)$ is piece-wise linear, we may search for the optimal point on one of its segments defined on $[a_1,a_2]$, such that $\mathcal{R}(a)=\alpha a + \beta$. The derivative of $\mathcal{R}(a)$ on $[a_1,a_2]$ is therefore:
    \begin{align}
        \frac{\partial V_a}{\partial a}(x) = x\gamma\frac{\alpha + \beta}{(1-a\gamma)^2}.
    \end{align}
    
    On the considered segment $\alpha+\beta$ is constant, and we may conclude that $V_a$ takes its minimal value either in $a_1$ or $a_2$. To minimize $V_a$ on the full domain of $a$, one just has to look at the singular points of the upper convex envelop, which are the coordinate $(\overline{g}_k,\overline{r}_k)$ of the arms $k$ that are on it with $\overline{g}_k<1$, which we concisely write $k\in\mathcal{P}_{<1}$. We call $V_{k^*}$ the value of the optimal constant policy playing repeatedly $g^* = \overline{g}_{k^*}$, with $r^*=\mathcal{R}(g^*)$. Consequently, $V_{g^*}$ may be written as follows:
    \begin{align}
        k^* &= \argmax_{k \in \mathcal{P}_{<1}} \frac{\overline{r}_k}{1- \gamma\overline{g}_k} \\
        V_{g^*}(x) &= x \frac{r^*}{1- \gamma g^*}.
    \end{align}
    
    Now we prove by contradiction that there is no possible policy improvement over $V_{g^*}$:
    \begin{align}
        0 < Q_{g^*}(x,g) - V_{g^*}(x) &= x\mathcal{R}(g) + \gamma V_{g^*}(\min(gx,x_{\mytop})) - x\frac{r^*}{1-\gamma g^*} \\
        &= x\mathcal{R}(g) + \min(gx,x_{\mytop})\gamma\frac{r^*}{1-\gamma g^*} - x\frac{r^*}{1-\gamma g^*} \\
        &= \frac{1}{1-\gamma g^*}\left(x\mathcal{R}(g) - x\gamma g^*\mathcal{R}(g) + \min(gx,x_{\mytop})\gamma r^* - xr^*\right),\label{eq:prop4-mainequation}
    \end{align}
    which has the sign of $\mathcal{R}(g) - \gamma g^*\mathcal{R}(g) + g\gamma r^* - r^*$ if $g<1$. It implies that:
    \begin{align}
        \frac{\mathcal{R}(g)}{1-\gamma g} \geq \frac{r^*}{1-\gamma g^*},
    \end{align}
    which is contradictory with the optimality in $k^*$. 
    
    If $g\geq 1$, we distinguish two cases: $\mathcal{R}(g^*)< 0$ and $\mathcal{R}(g^*)\geq 0$. If $\mathcal{R}(g^*)< 0$, we get:
    \begin{align}
        \frac{x\mathcal{R}(g) - g^*x\gamma\mathcal{R}(g) +  \min(gx,x_{\mytop})\gamma r^* - xr^*}{x} &\leq \mathcal{R}(g) - g^*\gamma\mathcal{R}(g) + \gamma r^* - r^* \nonumber\\
        &\leq (1-g^*\gamma)\mathcal{R}(g) - (1-\gamma)\mathcal{R}(g^*) \nonumber\\
        &\leq (1-g^*\gamma)\max_{g\geq 1} \mathcal{R}(g) - (1-\gamma)r^*, \nonumber
    \end{align}
    which is always negative when $\max_{g\geq 1} \mathcal{R}(g)\leq \mathcal{R}(g^*)$, or under the assumption we made on $\gamma$:
    \begin{align}
        \gamma \geq \max_{k\in\mathcal{P}_{< 1}} \frac{\max_{g\geq 1} \mathcal{R}(g) - \overline{r}_{k}}{\overline{g}_{k}\max_{g\geq 1} \mathcal{R}(g) - \overline{r}_{k}}.
    \end{align}
    This condition ensures that $\gamma$ is close enough to 1 so that the optimal policy would not be to keep up a population of $x$, and lose rewards, rather than investing to reduce the population.
    
    If $g\geq 1$ and $r^*\geq 0$, then we quickly consider the case where $g\leq \frac{1}{\gamma}$, and observe that, in this case $1-g^*\gamma\geq 0$ and $\mathcal{R}(g)< 0$, therefore the first term of Equation \ref{eq:prop4-mainequation} is negative, and $\min(gx,x_{\mytop})\gamma r^* - xr^*\leq xr^*(\gamma g - 1)$ is also negative. In the remaining case, when $g\geq \frac{1}{\gamma}$ and $r^*\geq 0$, let us upper bound the upper convex envelop by its local linear expression around $\frac{1}{\gamma}$: $\mathcal{R}(g')=\alpha g' + \beta$. We know by the concavity of the upper convex envelop that $r^*\leq \alpha g^* + \beta$, and $\mathcal{R}(g)\leq \alpha g + \beta$, which gives us:
    \begin{align}
        &\frac{x\mathcal{R}(g) - g^*x\gamma\mathcal{R}(g) +  \min(gx,x_{\mytop})\gamma r^* - xr^*}{x} \\
        &\qquad\qquad\qquad\qquad\qquad\leq \mathcal{R}(g) - g^*\gamma\mathcal{R}(g) + g\gamma r^* - r^* \\
        &\qquad\qquad\qquad\qquad\qquad\leq (1-g^*\gamma)(\alpha g + \beta) + (g\gamma-1)(\alpha g^* + \beta) \\
        &\qquad\qquad\qquad\qquad\qquad= \alpha g + \beta - \alpha g g^*\gamma - \beta g^*\gamma  + \alpha g g^*\gamma + \beta g\gamma -\alpha g^* - \beta\\
        &\qquad\qquad\qquad\qquad\qquad= (\alpha + \beta\gamma)(g-g^*),
    \end{align}
    which has to be non positive, because $g\geq 1 > g^*$ and $\alpha + \beta\gamma = \gamma\mathcal{R}(\frac{1}{\gamma}) \leq \gamma\max_{g\geq 1} \mathcal{R}(g) \leq 0$.
    
    We may therefore conclude that, under those conditions, $V_{g^*}$ is the optimal value, and constantly performing $g^*$ is an optimal policy, and since $g^*<1$, it will geometrically deplete the crowd.
\end{proof}

\begin{restatable}[PMDP-optimal value comparison]{corollary}{PMDPoptimal}
    $V^*_r \geq V^*_p$.
    \label{cor:PMDPoptimal}
\end{restatable}
\begin{proof}
  We use the concavity of the value functions demonstrated in Property \ref{prop:increasing} in conjunction with the Jensen's inequality:
  \begin{align}
      V^*_p(x) &= \sum_{k\in[K]}\left(\pi^*_k(x) \overline{r}_k\right) + \gamma \mathbb{E}_{x'\sim P_p(x,\pi^*(x))}V^*_p(x') \\
      &\leq \sum_{k\in[K]}\left(\pi^*_k(x) \overline{r}_k\right) + \gamma V^*_p(\mathbb{E}_{x'\sim P_p(x,\pi^*(x))}x') \\
      V^*_p(x_t) &\leq \sum_{k\in[K]}\left(\pi^*_k(x) \overline{r}_k\right) + \gamma V(x_{t+1}) \quad\quad\text{with } x_{t+1} = \mathbb{E}_{x'\sim P_p(x_t,\pi^*(x_t))}x' \\
      V^*_p(x_0) &\leq \sum_{t=0}^\infty \gamma^t \sum_{k\in[K]}\left(\pi^*_k(x_t) \overline{r}_k\right) \quad\quad\text{with } x_{t+1} = \mathbb{E}_{x'\sim P_p(x_t,\pi^*(x_t))}x' \\
      &= V^*_r(x_0)
  \end{align}
\end{proof}

\begin{restatable}[cross-optimal values comparison]{corollary}{crossoptimal}
    $V^{\mydiese}_r \geq V^*_p$.
    \label{cor:crossoptimal}
\end{restatable}
\begin{proof}
    By construction, we have $V^{\mydiese}_r(x) \geq V^*_r(x)$, and from Corollary \ref{cor:PMDPoptimal}, we have $ V^*_r(x)\geq V^*_p(x)$. This concludes the proof.
\end{proof}

\begin{restatable}[Pseudo-lipschitzness]{corollary}{lipschitzness}
    \begin{align}
        \forall x>x'\geq 0,\quad\quad V^{\mydiese}_r\left(x\right)-V^{\mydiese}_r\left(x'\right)&\leq \cfrac{x-x'}{x} V^{\mydiese}_r(x),
  \end{align}
  \label{cor:lipschitzness}
\end{restatable}
\begin{proof}
    This is a direct consequence of the concavity property demonstrated in Property \ref{prop:increasing} and the trivial fact that $V^{\mydiese}_r(0)=0$:
    \begin{align}
        \forall 0\leq \lambda \leq 1, \quad V^{\mydiese}_r(\lambda x) &\geq \lambda V^{\mydiese}_r(x) + (1-\lambda)V^{\mydiese}_r(0) \\
        &\geq \lambda V^{\mydiese}_r(x) \\
        \Leftrightarrow \quad\forall 0\leq x' \leq x, \quad V^{\mydiese}_r(x') &\geq \cfrac{x'}{x} V^{\mydiese}_r(x) \\
        \Leftrightarrow  \quad V^{\mydiese}_r(x)-V^{\mydiese}_r(x') &\leq V^{\mydiese}_r(x) - \cfrac{x'}{x} V^{\mydiese}_r(x) \\
        &= \cfrac{x-x'}{x} V^{\mydiese}_r(x),
    \end{align}
  which concludes the proof.
\end{proof}

In order to give a value to $\pi^*$ in the ROeMDP, we extend $\pi^*$ to the domain of definition of the ROeMDP as the interpolation to its closest integer values:
\begin{align}
    \pi^*(x) = (\lceil x \rceil-x)\pi^*(\lfloor x \rfloor) + (x-\lfloor x \rfloor)\pi^*(\lceil x \rceil).
\end{align}

\begin{restatable}[Optimal value error upper bound in the PMDP when $\max_{g\geq 1} \mathcal{R}(g) < 0$]{lemma}{OeMDPerrorbelow}
     \begin{align}
         \text{If } \:\max_{g\geq 1} \mathcal{R}(g) < 0,\quad\quad V^{\mydiese}_o(x) - V^{\mydiese}_p(x) &\leq e^{-s_0(x_{\mytop} - x_0)} \left(x_{\mytop}+g_{\mytop}\right) \max_{k\in[K], \text{ s.t. }\overline{g}_k < 1} \frac{\overline{r}_{k}}{1-\overline{g}_{k}},
     \end{align}
     where $s_0$ is a constant, as defined in Lemma \ref{lem:underthetop}.
    \label{lem:OeMDPerrorbelow}
\end{restatable}
\begin{proof}
    We know from Theorem \ref{th:below(1,0)} that $\pi^{\mydiese}$ is constantly selecting a single arm when $\max_{g\geq 1} \mathcal{R}(g) < 0$. As a consequence, and since there is no discounting ($\gamma=1$), we may consider that the batches are of size 1. Let $k_{\mydiese}$ be this arm, $\overline{g}_{\mydiese}$ its expected growth and $\overline{r}_{\mydiese}$ its expected reward (we set $\gamma=1$). Let $G_{x_0,x_{\mytop}}$ be the random variable of the sum of rewards $R_t$ collected during the process starting from crowd $x_0$ with maximal crowd $x_{\mytop}\geq x_0$. If $\overline{r}_{\mydiese}>0$, then we have\footnote{The inequalities are reversed if $\overline{r}_{\mydiese}<0$, and $V^{\mydiese}_o(x) - V^{\mydiese}_p(x) < 0$.}:
    \begin{align}
        V^{\mydiese}_o(x) - V^{\mydiese}_p(x) &= \mathbb{E}[G_{x_0,\infty}] - \mathbb{E}[G_{x_0,x_{\mytop}}] \\
        &= \mathbb{E}[G_{x_0,\infty} - G_{x_0,x_{\mytop}}] \\
        &= \mathbb{P}\left(\exists T_{\mytop}, \text{ s.t. } X_{T_{\mytop}} > x_{\mytop} | X_0=x_0\right)\mathbb{E}\left[\sum_{t=0}^{T_{\mytop}-1} R_{t} + G_{X_{T_{\mytop}},\infty} - \sum_{t=0}^{T_{\mytop}-1} R_{t} + G_{x_{\mytop},x_{\mytop}}\right]\nonumber \\
        &= \underbrace{\mathbb{P}\left(\exists T_{\mytop}, \text{ s.t. } X_{T_{\mytop}} > x_{\mytop} | X_0=x_0\right)}_{\text{Lemma \ref{lem:underthetop}}}\bigg(\underbrace{\mathbb{E}\left[G_{X_{T_{\mytop}},\infty}\right]}_{\text{Theorem \ref{th:below(1,0)}}} - \underbrace{\mathbb{E}\left[G_{x_{\mytop},x_{\mytop}}\right]}_{\text{greater than 0}}\bigg) \\
        &\leq e^{-s_0(x_{\mytop} - x_0)} \underbrace{\mathbb{E}[X_{T_{\mytop}}]}_{\text{smaller than }x_{\mytop}+g_{\mytop}\text{ because batches of 1}} \frac{\overline{r}_{\mydiese}}{1-\overline{g}_{\mydiese}} \\
        &\leq e^{-s_0(x_{\mytop} - x_0)} \left(x_{\mytop}+g_{\mytop}\right) \frac{\overline{r}_{\mydiese}}{1-\overline{g}_{\mydiese}},
    \end{align}
    which concludes the proof.
\end{proof}

\begin{lemma}[Probability to exceed crowd under decreasing regime] Let $(\xi_{t,i})_{t\ge 0, i\ge 1}$ be a family of iid copies of a random variable $\xi$ taking values in $\mathbb N$, and not concentrated  on $\{0,1\}$. Let $(X_t)_{t\ge 0}$ be such that
\[X_{t+1}=\sum_{i=1}^{X_t}\xi_{t,i}\,.\]
Suppose that ${\mathbb E}[\xi]=m<1$ and that ${\mathbb E}[e^{s\xi}]<\infty$ for every $s\ge 0$. 
Then, there exists a unique $s_0 > 0$ such that $\mathbb{E}[e^{s_0\xi}]=e^{s_0}$, and we have
\[\mathbb{P}(\exists t, \text{ s.t. } X_t > x_{\mytop} | X_0=x_0) \leq e^{-s_0(x_{\mytop} - x_0)}\,.\]

In particular, if $\xi$ follows a geometric distribution, which is a commonly used law for modeling propagation of disease/information, we find that $s_0=-\ln{m}$, and therefore that:
\[\mathbb{P}(\exists t, \text{ s.t. } X_t > x_{\mytop} | X_0=x_0) \leq m^{x_{\mytop} - x_0}\,.\]
\label{lem:underthetop}
\end{lemma}
\begin{proof}
Let $\Lambda$ denote the cumulant generating function of $\xi$, that is $\Lambda(s) = \ln \mathbb{E}[e^{s\xi }]$, which exists and is finite for each $s\ge 0$ by assumption. Then $\Lambda(0)=0$, $\Lambda$ is continuous and convex. Note also that a Taylor expansion at $s\to0$ yields $\Lambda'(0)={\mathbb E}[\xi]=m$. Furthermore, since ${\mathbb P}(\xi\ge 2)\ge \epsilon>0$, we have  $\Lambda(s)\ge \ln(\epsilon e^{2s})$ and therefore
\begin{align*}
\Delta(s)-s 
\geq \ln\left(\epsilon e^{2s}\right)-s 
= s + \ln \epsilon,
\end{align*}
which implies that $\Delta(s)-s \to + \infty$ as $s\to\infty$. It follows that the equation $\Lambda(s)=s$ has aside from the trivial solution $s=0$, a  unique positive solution $s_0$, as claimed. 

In particular, we have ${\mathbb E}[e^{s_0 \xi}]=e^{s_0}$, which allows us to construct a martingale as follows. 
For $t \in \mathbb{N}$, let $M_t=e^{s_0X_t}$. Then, for each $t$, $M_t$ is integrable and
\begin{align*}
    \mathbb{E}[M_{t+1}|X_0,\dots, X_t] = \mathbb{E}\left[e^{\sum_{k=0}^{X_t}s_0\xi_{t,k}}\middle|X_0,\dots, X_t\right] 
    = \prod_{k=1}^{X_t}\mathbb{E}[e^{s_0\xi_{t,k}}] 
    = \prod_{k=1}^{X_t}e^{s_0} 
    \textbf{}= M_t\,,
\end{align*}
so that $(M_t)_{t\ge 0}$ is a martingale. 

Let now $T_{\mytop} = \inf\{ n, \text{ s.t. } X_n > x_{\mytop}\}$. Then $T_{\mytop}$ is a stopping time, so the stopped process $(M_{\min(T_{\mytop},t)})_{t \geq 0}$ is also a martingale. On the one hand, we have $\mathbb{E}[M_{\min(T_{\mytop},0)}]=e^{s_0x_0}$, and on the other hand, by the martingale property, for any $t \geq 0$:
\begin{align}
\label{optional_stopping}
    e^{s_0 x_0}=\mathbb{E}[M_{\min(T_{\mytop},0)}] &= \mathbb{E}[M_{\min(T_{\mytop},t)}] \notag\\
    &= \mathbb{E}[M_{\min(T_{\mytop},t)}\mathbbm{1}(T_{\mytop} = +\infty)] +  \mathbb{E}[M_{\min(T_{\mytop},t)}\mathbbm{1}(T_{\mytop} < +\infty)].
\end{align}
The process $M_{\min(T_{\mytop}, t)}$ is bounded (by $e^{s_0 x_{\mytop}}$) and therefore, since $X_t\to 0$ almost surely as $t\to\infty$ since the branching process is subcritical (${\mathbb E}[\xi]=m<1$, see \cite{AtNe1972}), it follows that
\begin{align}
\mathbb{E}[M_{\min(T_{\mytop},t)}\mathbbm{1}(T_{\mytop} = +\infty)] = \mathbb{E}[M_{t}\mathbbm{1}(T_{\mytop} = +\infty)] \xrightarrow[t\to\infty]{} \mathbb{P}(T_{\mytop} = +\infty).
\end{align}

Moreover since $M_{\min(T_{\mytop},t)}= e^{s_0 X_{T_{\mytop}}}$ for $t\ge T_{\mytop}$, we have
\begin{align}
\lim_{t\to +\infty}\mathbb{E}[M_{\min(T_{\mytop},t)}\mathbbm{1}(T_{\mytop} < +\infty)] \geq \mathbb{P}(T_{\mytop} < +\infty)e^{s_0x_{\mytop}}.
\end{align}

Injecting the above two inequalities in \eqref{optional_stopping} yields:
\begin{align}
\mathbb{P}_{x_0}(\exists t: X_t>x_{\mytop})=\mathbb{P}(T_{\mytop} < +\infty) \leq \frac{e^{s_0x_0}-1}{e^{s_0x_{\mytop}}-1}\le e^{s_0(x_0-x_{\mytop})},
\end{align}
since $1\le x_0\le x_{\mytop}$.
\end{proof}

    

\begin{restatable}[Value error upper bound in the OeMDP when $\max_{g\geq 1} \mathcal{R}(g) > 0$]{lemma}{OeMDPerrorabove}
    Let $\pi$ be a policy such that $V^{\pi}_o(x)$ increases with $x\in[x_{\mytop}]$, then, if $\max_{g\geq 1} \mathcal{R}(g) > 0$, we have the following upper bound on the error:
    \begin{align}
        V^{\pi}_o(x) - V^{\pi}_p(x) &\leq \cfrac{\gamma V^{\pi}_{\mytop}}{1-\gamma} \left(\frac{\gamma\zeta^x}{(1-\gamma)^2} + \cfrac{2+\dot{g}_{\mytop}\sqrt{\ln (x)}}{2\sqrt{x}}\right). \nonumber
    \end{align}
    \label{lem:OeMDPerrorabove}
\end{restatable}
\begin{proof}
    \begin{align}
        V^{\pi}_o(x) - V^{\pi}_p(x) &= \gamma \left(V^{\pi}_o P_o - V^{\pi}_p P_p\right)\pi(x)\\
         &= \gamma \left(V^{\pi}_o \left(P_o - P_p\right) + \left(V^{\pi}_o - V^{\pi}_p\right) P_p\right)\pi(x) \\
         &= \gamma V^{\pi}_o \left(P_o - P_p\right) \pi \left(\mathbb{I} - \gamma P_p \pi \right)^{-1}(x) \label{eq:inverse1}\\
         &= \frac{\gamma}{1-\gamma} \sum_{x'=0}^{x_{\mytop}} d^{\pi}_p(x,x') V^{\pi}_o \left(P_o - P_p\right) \pi(x'), \label{eq:dist_int}
    \end{align}
    where, for brevity, $\pi$ above returns the state-action couple resulting from the application of the policy to a given state, where $d^{\pi}_p(x,\cdot)$ is the normalized discounted sum of visited states, starting from $x$, under policy $\pi$ in the real stochastic environment. Line \ref{eq:inverse1} is obtained by moving the right-hand term to the left side of the equality and then the terms are factorized with $V^{\pi}_o - V^{\pi}_p$, and inverted. $\mathbb{I} - \gamma P_p \pi$ is always invertible because $\gamma<1$. Line \ref{eq:dist_int} is simply a rewriting of $\left(\mathbb{I} - \gamma P_p \pi \right)^{-1}$ which sums to $\frac{1}{1-\gamma}$ with the discounted visitation density $d^{\pi}_p(x,\cdot)$, which sums to 1. 
    
    Now, we are interested in estimating an upper bound of the term inside the sum:
    \begin{align}
         V^{\pi}_o \left(P_o - P_p\right)(x,a) &= \mathbb{E}_{x'_p \sim P_p(x,a,\cdot)}\left[V^{\pi}_o(x'_o) - V^{\pi}_o(x'_p)\right] \\
         &= \sum_{x'_p=0}^{x_{\mytop}} P_p(x,a,x'_p)\left(V^{\pi}_o(x'_o) - V^{\pi}_o(x'_p)\right),
    \end{align}
    where $x'_o$ is the deterministic successor of $x$ after executing $a$ in the OeMDP, and where $x'_p \sim P_p(x,a,\cdot)$ is the stochastic successor of $x$ after executing $a$ in the PMDP. Since the value is monotonically increasing with $x\in[x_{\mytop}]$ (by assumption), we may upper bound the error on the transitions that are under $x'_o$\footnote{For the sake of simplicity, we do not deal with the rounding errors. $\sum_{i=x}^{y}$ with $x$ and $y$ real numbers will mean the sum for all $i\in \mathbb{N}\cup[x,y]$.}:
    \begin{align}
         V^{\pi}_o \left(P_o - P_p\right)(x,a) &\leq \sum_{x'_p=0}^{x'_o} P_p(x,a,x'_p)\left(V^{\pi}_o(x'_o) - V^{\pi}_o(x'_p)\right) \\
         &= \sum_{x'_p=0}^{x'_o-CI_1} P_p(x,a,x'_p)\left(V^{\pi}_o(x'_o) - V^{\pi}_o(x'_p)\right) + \sum_{x'_p=x'_o-CI_1}^{x'_o} P_p(x,a,x'_p)\left(V^{\pi}_o(x'_o) - V^{\pi}_o(x'_p)\right) \label{eq:intdecomposition}\\
         &\leq V^{\pi}_o(x'_o-CI_1)\exp\left(-2 \frac{CI_1^2}{x \dot{g}_{\mytop}^2}\right) + \sum_{x'_p=x'_o-CI_1}^{x'_o} P_p(x,a,x'_p)(x'_o-x'_p)\cfrac{V^{\pi}_o(x'_o)}{x'_o} \label{eq:hoeffding}\\
         &\leq V^{\pi}_o(x'_o)\exp\left(-2 \frac{CI_1^2}{x \dot{g}_{\mytop}^2}\right) + CI_1\cfrac{V^{\pi}_o(x'_o)}{x'_o} \sum_{x'_p=x'_o-CI_1}^{x'_o} P_p(x,a,x'_p) \label{eq:boundbyCI}\\
         &\leq V^{\pi}_o(x'_o)\exp\left(-2 \frac{CI_1^2}{x \dot{g}_{\mytop}^2}\right) + CI_1\cfrac{V^{\pi}_o(x'_o)}{x'_o} \label{eq:lessthan1} \\
         &\leq V^{\pi}_o(x'_o)\left(\exp\left(-2 \frac{CI_1^2}{x \dot{g}_{\mytop}^2}\right) + \cfrac{CI_1}{x}\right), \label{eq:factorization}
    \end{align}
    where line \ref{eq:intdecomposition} is obtained by decomposing the sum in parts at a cutting point $x'_o - CI_1$ that is going to be determined later. Line \ref{eq:hoeffding} is obtained by applying Hoeffding's bound on the first sum and Corollary \ref{cor:lipschitzness} on the second term. Line \ref{eq:boundbyCI} is obtained by upper bounding $x'_o-x'_p$ with $CI_1$. Line \ref{eq:lessthan1} is obtained because the transition kernel sums to 1. Finally, Line \ref{eq:factorization} is a simple factorization and a lower bound of $x'_o$ as $x$, since we assumed that $\mathcal{R}(\frac{1}{\gamma})>0$, and Property \ref{prop:existence} states that $\pi(x)\geq 1$ for all $x$ under this assumption. In particular, if we choose $CI_1 = \frac{\dot{g}_{\mytop}}{2}\sqrt{x\ln x}$, we get:
    \begin{align}
        V^{\pi}_o \left(P_o - P_p\right)(x,a) &\leq V^{\pi}_o(x'_o)\left(\exp\left(-\frac{\ln x}{2}\right) + \cfrac{\dot{g}_{\mytop}\sqrt{\ln x}}{2\sqrt{x}}\right) \\
        &= V^{\pi}_o(x'_o)\left(\frac{1}{\sqrt{x}} + \cfrac{\dot{g}_{\mytop}\sqrt{\ln x}}{2\sqrt{x}}\right) \label{eq:prelimresult}
    \end{align} 
    
    Starting back from Equation \ref{eq:dist_int}:
    \begin{align}
        V^{\pi}_o(x) - V^{\pi}_p(x) &= \frac{\gamma}{1-\gamma} \int_{\mathcal{X}_p} d^{\pi}_p(x,x') V^{\pi}_o \left(P_o - P_p\right) (x',\pi(x'))dx' \\
        &= \frac{\gamma}{1-\gamma} \sum_{x'=0}^{x-1} d^{\pi}_p(x,x') V^{\pi}_o \left(P_o - P_p\right) (x',\pi(x')) + \frac{\gamma}{1-\gamma} \sum_{x'=x}^{x'_{\mytop}} d^{\pi}_p(x,x') V^{\pi}_o \left(P_o - P_p\right) (x',\pi(x'))' \label{eq:intdecomposition2}\\
        &\leq \frac{\gamma}{1-\gamma} V^{\pi}_{\mytop}\frac{\gamma\zeta^x}{(1-\gamma)^2} 
        + \frac{\gamma}{1-\gamma} \sum_{x'=x}^{x_{\mytop}} d^{\pi}_p(x,x') V^{\pi}_o(x')\left(\frac{1}{\sqrt{x'}} + \cfrac{\dot{g}_{\mytop}\sqrt{\ln x'}}{2\sqrt{x'}}\right) \label{eq:hoeffding2} \\
        &\leq \cfrac{\gamma V^{\pi}_{\mytop}}{1-\gamma} \frac{\gamma\zeta^x}{(1-\gamma)^2}
        + \cfrac{\gamma V^{\pi}_{\mytop}}{1-\gamma} \left(\frac{1}{\sqrt{x}} + \cfrac{\dot{g}_{\mytop}\sqrt{\ln x}}{2\sqrt{x}}\right) \sum_{x'=x}^{x_{\mytop}} d^{\pi}_p(x,x') \label{eq:x'replacement}\\
        &\leq \cfrac{\gamma V^{\pi}_{\mytop}}{1-\gamma} \left(\frac{\gamma\zeta^x}{(1-\gamma)^2} + \cfrac{2+\dot{g}_{\mytop}\sqrt{\ln (x)}}{2\sqrt{x}}\right), \label{eq:factorization2}
    \end{align}
    where line \ref{eq:intdecomposition2} is once more a decomposition of the sum in two parts. Line \ref{eq:hoeffding2} replaces the result of Lemma \ref{lem:discountedtimeunderthreshold} with constant $\zeta<1$, and by injecting the result of Equation \ref{eq:prelimresult} inside the second sum. In line \ref{eq:x'replacement}, we upper bound the expression by replacing $x'$ with the value that maximizes it. Finally, line \ref{eq:factorization2} is a simple refactorization that concludes the proof.  See below for details about $\zeta$.
    
    $\zeta$ is a constant related to the problem and the policy resulting from solving the ROeMDP associated with it. The choice of $x'_{\mytop} \leq x_{\mytop}$ used in Lemma \ref{lem:discountedtimeunderthreshold} is balance between choosing it high and such that $m=\min_{x\in[x'_{\mytop}]}\pi^{\mydiese}(x)$ is large\footnote{Actually $m$ is the expectation the minimum over the random variables, and not their mean.}:
    \begin{align}
        &\sum_{x'=0}^{x-1} d^{\pi}_p(x,x') = \sum_{t=0}^\infty \gamma^t \mathbb{P}(X_t < x_{\mybot}|\pi = \pi^{\mydiese}) \\
        &\leq \sum_{t=0}^\infty \gamma^t \mathbb{P}(X_t < x_{\mybot}|\pi = m) \\
        &\leq \min_{x'_{\mytop}\in[x,x_{\mytop}]} \left\{ \exp\left(- \frac{x (\pi(x'_{\mytop})-1)^3}{2\pi(x'_{\mytop})(\pi(x'_{\mytop})(\pi(x'_{\mytop})-1)+\sigma^2)}\right) +  \exp\left(-\frac{x'_{\mytop} (\pi(x'_{\mytop})-1)^2}{4(\sigma^2+\pi(x'_{\mytop})^2)}\right) \right\} \\
        &\leq \min_{x'_{\mytop}\in[x,x_{\mytop}]} \left\{ \exp\left(- \frac{x (\pi(x'_{\mytop})-1)^3}{2\pi(x'_{\mytop})(\pi(x'_{\mytop})(\pi(x'_{\mytop})-1)+\sigma^2)}\right) +  \exp\left(-\frac{x (\pi(x'_{\mytop})-1)^2}{4(\sigma^2+\pi(x'_{\mytop})^2)}\right) \right\} \\
        &\leq \min_{x'_{\mytop}\in[x,x_{\mytop}]} \max\left\{ \exp\left(- \frac{(\pi(x'_{\mytop})-1)^3}{2\pi(x'_{\mytop})(\pi(x'_{\mytop})(\pi(x'_{\mytop})-1)+\sigma^2)}\right)\; ;\;  \exp\left(-\frac{ (\pi(x'_{\mytop})-1)^2}{4(\sigma^2+\pi(x'_{\mytop})^2)}\right) \right\}^{x} \nonumber\\
        &\leq \zeta^{x}
    \end{align}
    If $x$ gets too close from $x_{\mytop}$, it may happen that $x'_{\mytop}$ gets constrained by it being larger than $x$. In this case, one may choose $x'_{\mytop}$ smaller and replace $x$ with $x'_{\mytop}$ in the exponentiation of $\zeta$. This detail is omitted in the main result for the sake of conciseness.
\end{proof}

\begin{lemma}[Discounted time under threshold]
    We consider the following process:
    \begin{align}
        \left\{\begin{array}{rl}
             X_0 &= x_0  \\
             X_{t+1} &= \left\{\sum_{i=1}^{X_t} \xi_{t,i}\right\} \wedge x_{\mytop} \,,
        \end{array}\right.
    \end{align}
    where $(\xi_{t,i})_{t\ge 0,i\ge 1}$ are iid copies of a random variable $\xi$ with expected value $\mathbb{E}[\xi]=m>1$ and finite variance $\sigma^2$. 
    Then, for any $x_{\mybot}\in \{0,\dots, x_0\}$we have
    \begin{align}\label{eq:claimed_bound}
    \sum_{t=0}^\infty \gamma^t \mathbb{P}(X_t < x_{\mybot}) \leq  \frac{\gamma}{(1-\gamma)^2}\left\{ \exp\left(- \frac{x_0 (m-1)^3}{2m(m(m-1)+\sigma^2)}\right) +  \exp\left(-\frac{x_{\mytop} (m-1)^2}{4(\sigma^2+m^2)}\right) \right\}.\nonumber
    \end{align}
    \label{lem:discountedtimeunderthreshold}
\end{lemma}


\begin{proof}The following version of the process where the upperbound has been dropped will be useful:
    \begin{align}
        \left\{\begin{array}{rl}
             X'_0 &= x_0  \\
             X'_{t+1} &= \left\{\sum_{i=1}^{X'_t} \xi_{t,i}\right\} \quad\text{with } \mathbb{E}[\xi]=m>1
        \end{array}\right.
    \end{align}
    One can couple $(X_t)$ and $(X'_t)$ in such a way that that $X_t=X'_t$ as long as $X'_t\le x_{\mytop}$, that is for every  $t\leq T\coloneqq \inf\left\{t'\ge 0 :X'_{t'}>x_{\mytop}\right\}$. In the following, we write  $\mathbb{P}_x(\cdot) = \mathbb{P}(\cdot|X_0=x)$.\textbf{}
Straightforward computation and the strong Markov property used at time $T$ yields
    \begin{align}
        \sum_{t=0}^\infty \gamma^t \mathbb{P}_{x_0}(X_t < x_{\mybot}) &= \sum_{t=0}^\infty \mathbb{E}_{x_0}\left[\gamma^t \mathds{1}_{X_t < x_{\mybot}}\right] \\
        &=  \mathbb{E}_{x_0}\left[\sum_{t=0}^\infty\gamma^t \mathds{1}_{X_t < x_{\mybot}}\right] \\
        &= \mathbb{E}_{x_0}\left[\sum_{t=0}^{T-1}\gamma^t \mathds{1}_{X_t < x_{\mybot}}+\sum_{t=T}^\infty\gamma^t \mathds{1}_{X_t < x_{\mybot}}\right] \\
        &= \mathbb{E}_{x_0}\left[\sum_{t=0}^{T-1}\gamma^t \mathds{1}_{X'_t < x_{\mybot}}+\sum_{t=T}^\infty\gamma^t \mathds{1}_{X_t < x_{\mybot}}\right] \\
        &\leq \mathbb{E}_{x_0}\left[\sum_{t=0}^{\infty}\gamma^t \mathds{1}_{X'_t < x_{\mybot}}\right]+\mathbb{E}_{x_0}\left[\sum_{t=0}^\infty\gamma^{t+T} \mathds{1}_{X_{t+T} < x_{\mybot}}\right] \\
        &\leq  \sum_{t=0}^\infty \gamma^t \mathbb{P}_{x_0}(X'_t < x_{\mybot})+\mathbb{E}_{x_0}\left[\gamma^T\right]\mathbb{E}_{x_{\mytop}}\left[\sum_{t=0}^\infty\gamma^{t} \mathds{1}_{X_t < x_{\mybot}}\right]\\
        &\leq  \sum_{t=0}^\infty \gamma^t \mathbb{P}_{x_0}(X'_t < x_{\mybot})+\mathbb{E}_{x_{\mytop}}\left[\sum_{t=0}^\infty\gamma^{t} \mathds{1}_{X_t < x_{\mybot}}\right] \label{eq:two_terms}\,.
    \end{align}
    
    First, we deal with the first term where we have a sum of $x_0$ independent random variables $Z_t$ defined as follows:
    \begin{align}
        \left\{\begin{array}{rl}
             Z_0 &= 1  \\
             Z_{t+1} &= \sum_{i=1}^{Z_t} \xi_{t,i}\,.
        \end{array}\right.
    \end{align}
    In particular, for each $t\geq 1$, we have $\mathbb{E}Z_t=m\mathbb{E}Z_{t-1}=m^t$, and $Z_t\geq 0$. Then, writing $(Z^{(j)}_\bullet)_{1\le j\le x_0}$ for $x_0$ iid copies of $(Z_t)_{t\ge 0}$, we have
    \begin{align}
        X'_t &= \sum_{j=1}^{x_0} Z^{(j)}_t\,.
    \end{align}
    
    For summands distributed like $Z_t$, we have one-sided Bernstein concentration inequality:
    \begin{align}
        \mathbb{P}_{x_0}(X'_t < x_{\mybot}) &= \mathbb{P}\left(\sum_{j=1}^{x_0} Z^{(j)}_t < x_{\mybot}\right) \\
        &= \mathbb{P}\left(\sum_{j=1}^{x_0}\left\{ Z^{(j)}_t - \mathbb{E}Z_t\right\} < x_{\mybot}-x_0 m^t\right) \\
        &\leq \exp\left(-\frac{\left(x_0 m^t-x_{\mybot}\right)^2}{2 x_0 v_t}\right),
    \end{align}
    where $v_t = \mathbb{E}[Z_t^2]=\mathrm{Var}(Z_t)+\mathbb{E}[Z_t]^2$. It is standard \citep{AtNe1972} that, when $m\ne 1$ and writing $\sigma^2=\mathrm{Var}(\xi)$, we have
    \begin{align}\label{eq:vt}
        v_t &= \frac{\sigma^2m^{t-1}(m^t-1)}{m-1} + m^{2t}\\
        &=m^{2t} \left(1 + \frac{\sigma^2}{m(m-1)}(1-m^{-t})\right) \\
        &\le m^{2t} \left(1+\frac{\sigma^2}{m(m-1)}\right)\,.
    \end{align}
    It follows that 
    \begin{align}\label{eq:decomp_bound}
        \mathbb{P}_{x_0}(X'_t<x_{\mybot}) 
        &\le \exp\left( - \frac{(x_0(m^t-1)+x_0-x_{\mybot})^2}{2x_0v_t}\right) \\
        &= \exp\left(- x_0\frac{ (m^t-1)^2}{2v_t}\right) \times \exp\left(-\frac{(m^t-1)(x_0-x_{\mybot})}{v_t}\right) \times \exp\left(-\frac{(x_0-x_{\mybot})^2}{2x_0v_t}\right)\,.\nonumber
    \end{align}
    Note that, the bound in \eqref{eq:vt} implies that 
    \begin{align}
        \exp\left(-x_0 \frac{(m^t-1)^2}{2v_t}\right) 
        &\le \exp\left(-x_0 \frac{m(m-1)(1-m^{-t})^2}{2(m(m-1)+\sigma^2)}\right) \\
        &\le \exp\left(-x_0 \frac{(m-1)^3}{2m(m(m-1)+\sigma^2)}\right)\,,\label{eq:exp_drift}
    \end{align}
    for all $t\ge 1$, where the second line is obtained because $1-m^{-t}>1-m^{-1}=\frac{m-1}{m}$. 
    The second and third factors in \eqref{eq:decomp_bound} are smaller than 1 and dropped. We deduce that 
    \begin{align}
        \sum_{t\ge 0} \gamma^t \mathbb{P}_{x_0}(X'_t<x_{\mybot}) \le \frac{\gamma}{1-\gamma} \exp\left(-x_0 \frac{(m-1)^3}{2m(m(m-1)+\sigma^2)}\right)\,.\label{eq:first_term}
    \end{align}
    
    We now move on to the bound on the second term of \eqref{eq:two_terms}. In order to deal with the push down resulting from the upper bound at $x_{\mytop}$, we proceed as follows. We assume here for simplicity that $x_{\mytop}>2x_{\mybot}$; otherwise, we always have $x_{\mytop}>\kappa x_{\mybot}$ for some $\kappa>1$ and we replace 2 by $\kappa$ in the following definitions. Let $x_m:=\lceil x_{\mytop}/2\rceil$. Define the auxiliary process $X''_0=X_0$ and 
    \begin{align}
        X''_{t+1} = \sum_{i=1}^{x_m} \xi_{t,i}.
    \end{align}
    Let $T_m:=\inf\{t\ge 0: X''_t<x_m\}$. Then, since dropping some individuals only decreases the population, for each $t<T_m$, we have $X_t\ge X''_t \wedge x_{\mytop}$. Furthermore, the random variables $(X''_t)_{t\ge 1}$ are actually independent and identically distributed. Note that
    \begin{align}
        \mathbb{P}(X_t<x_{\mybot}) 
        &\le \mathbb{P}(X_t<x_m) \\
        &\le \mathbb{P}(X_t<x_m, X''_k\ge x_m \forall k\le t) + \mathbb{P}(\exists k\le t: X''_k < x_m )\\
        &\le \mathbb{P}\left(\exists k\le t: X''_k < x_m \right).
    \end{align}
    It follows easily by the union bound and Bernstein's one-sided inequality that
    \begin{align}
        \mathbb{P}(X_t<x_{\mybot})
        & \le t \cdot \mathbb{P}\left(\sum_{i=1}^{x_m} \xi_i<x_m\right)\\
        & = t \cdot \mathbb{P}\left(\sum_{i=1}^{x_m} (\xi_i-m) < x_m (1-m)\right)\\
        & \le t\exp\left(-\frac{x_m^2 (m-1)^2}{2x_m v}\right)\\
        & = t \exp\left(-\frac{x_m (m-1)^2}{2v}\right),
    \end{align}
    where $v=\mathbb{E}[\xi^2]=\sigma^2+m^2$.
    The second term of the right-hand side of \eqref{eq:two_terms} is therefore such that
    \begin{align}\label{eq:second_term}
        \mathbb{E}_{x_{\mytop}} \left[\sum_{t\ge 0} \gamma^t \mathds{1}_{X_t<x_{\mybot}}\right]
        &= \sum_{t\ge 0} \gamma^t \mathbb{P}(X_t<x_{\mybot})\\
        &\le \exp\left(-\frac{x_m (m-1)^2}{2v}\right) \sum_{t\ge 0} t \gamma^t  \\
        & = \frac{\gamma}{(1-\gamma)^2} \exp\left(-\frac{x_m (m-1)^2}{2v}\right)\,.
    \end{align}
    Putting \eqref{eq:first_term} and \eqref{eq:second_term} together and rejoining the expressions for $v_t$ and $v$ yields \eqref{eq:claimed_bound}.
\end{proof}

\begin{lemma}[Upper convex envelop reward divergence]
    Let $\mathcal{R}_1$ and $\mathcal{R}_2$ be two reduced reward functions defined on the same interval $[\overline{g}_{\mybot},\overline{g}_{\mytop}]$. Let $V^\pi_1$ and $V^\pi_2$ be the values of some policy $\pi$ in the ROeMDPs respectively induced by $\mathcal{R}_1$ and $\mathcal{R}_2$. Then, we have:
    \begin{align}
       \big\lVert V^\pi_1 - V^\pi_2\big\rVert_\infty \leq \frac{x_{\mytop}}{1-\gamma}\big\lVert\mathcal{R}_1 - \mathcal{R}_2\big\rVert_\infty
    \end{align}
    \label{lem:rewardpareto}
\end{lemma}
\begin{proof}
    The dynamics are not affected by the upper convex envelop reward divergence: $\forall t, x_{t+1} = \min(x_t\pi(x_t),x_{\mytop}))$. As a consequence, the value error is the discounted sum of errors made on the rewards:
    \begin{align}
        V^\pi_1(x_0) - V^\pi_2(x_0) &= x_0\mathcal{R}_1(\pi(x_0)) - x_0\mathcal{R}_2(\pi(x_0)) + \gamma V^\pi_1(x_1) - \gamma V^\pi_2(x_1) \\
        &= \sum_{t=0}^{\infty} x_t \gamma^t \left(\mathcal{R}_1(\pi(x_t)) - \mathcal{R}_2(\pi(x_t))\right) \\
        &\leq \sum_{t=0}^{\infty} x_{\mytop} \gamma^t \big\lVert\mathcal{R}_1 - \mathcal{R}_2\big\rVert_\infty \\
        &= \frac{x_{\mytop}}{1-\gamma}\big\lVert\mathcal{R}_1 - \mathcal{R}_2\big\rVert_\infty
    \end{align}
\end{proof}

\begin{lemma}[Upper convex envelop domain divergence]
    Let $\mathcal{R}$ be a reduced reward function defined on $[\overline{g}_{\mybot},\overline{g}_{\mytop}]$. Let $M$ be the ROeMDP induced by $\mathcal{R}$ on action set $\mathcal{G} = [\overline{g}_{\mybot},\overline{g}_{\mytop}]$ and $\widehat{M}$ be the ROeMDP induced by $\mathcal{R}$ on action set $\widehat{\mathcal{G}} = [\widehat{g}_{\mybot},\widehat{g}_{\mytop}] \subset \mathcal{G}$. Let $V^{\mydiese}_r$ and $\widehat{V}^{\mydiese}_r$ be the respective optimal values in $M$ and $\widehat{M}$. Then, we have the optimal value error $V^{\mydiese}_r(x) - \widehat{V}^{\mydiese}_r(x)$ that decreases linearly with upper convex envelop domain divergence: $\widehat{g}_{\mybot}-\overline{g}_{\mybot}$ and $\overline{g}_{\mytop}-\widehat{g}_{\mytop}$.
    \label{lem:domainpareto}
\end{lemma}
\begin{proof}
    Note that, by convexity assumption, we know that:
    \begin{align}
        \mathcal{R}(\overline{g}_{\mybot})- \mathcal{R}(\widehat{g}_{\mybot}) &\leq \alpha_{\mybot}(\overline{g}_{\mybot} - \widehat{g}_{\mybot}) \quad\text{where } \mathcal{R}(g) = \alpha_{\mybot} g + \beta_{\mybot} \text{ in the upper vicinity of }\widehat{g}_{\mybot}\\
        \mathcal{R}(\overline{g}_{\mytop})- \mathcal{R}(\widehat{g}_{\mytop}) &\leq \alpha_{\mytop}(\overline{g}_{\mytop} - \widehat{g}_{\mytop})
         \quad\text{where } \mathcal{R}(g) = \alpha_{\mytop} g + \beta_{\mytop} \text{ in the lower vicinity of }\widehat{g}_{\mytop}
    \end{align}
    
    We split the proof in two cases: \textbf{(I)} when $\max_{g\geq 1} \mathcal{R}(g)\leq 0$, and \textbf{(II)} when $\max_{g\geq 1} \mathcal{R}(g)> 0$. 
    
    \textbf{(I)} When $\max_{g\geq 1} \mathcal{R}(g)\leq 0$, according to Theorem \ref{th:below(1,0)}, the optimal values are:
    \begin{align}
        V^{\mydiese}_r(x) &= x \frac{\mathcal{R}(g_*)}{1- \gamma g_*} \quad\text{with } g_* = \argmax_{g \in \mathcal{G}\cap [0,1)} \frac{\mathcal{R}(g)}{1- \gamma g}  \\
        \widehat{V}^{\mydiese}_r(x) &= x \max_{g \in \widehat{\mathcal{G}}\cap [0,1)} \frac{\mathcal{R}(g)}{1- \gamma g} \\
        V^{\mydiese}_r(x) - \widehat{V}^{\mydiese}_r(x) &\leq x \max \left\{
        \underbrace{\frac{\mathcal{R}(\overline{g}_{\mybot})}{1- \gamma \overline{g}_{\mybot}} - \frac{\mathcal{R}(\widehat{g}_{\mybot})}{1- \gamma \widehat{g}_{\mybot}}}_{\text{case: } g_*<\widehat{g}_{\mybot}},
        \underbrace{0}_{\text{case: } g_*\in\widehat{\mathcal{G}}},
        \underbrace{\frac{\mathcal{R}(\overline{g}_{\mytop})}{1- \gamma \overline{g}_{\mytop}} - \frac{\mathcal{R}(\widehat{g}_{\mytop})}{1- \gamma \widehat{g}_{\mytop}}}_{\text{case: } g_*>\widehat{g}_{\mytop}}
        \right\} \label{eq:max}
    \end{align}
    
    Below, we unfold for the case $g_*>\widehat{g}_{\mytop}$, but the same may be identically done for the case $g_*<\widehat{g}_{\mybot}$:
    \begin{align}
        \frac{\mathcal{R}(\overline{g}_{\mytop})}{1- \gamma \overline{g}_{\mytop}} - \frac{\mathcal{R}(\widehat{g}_{\mytop})}{1- \gamma \widehat{g}_{\mytop}} &=\frac{(1- \gamma \widehat{g}_{\mytop})\mathcal{R}(\overline{g}_{\mytop})- (1- \gamma \overline{g}_{\mytop})\mathcal{R}(\widehat{g}_{\mytop})}{(1- \gamma \overline{g}_{\mytop})(1- \gamma \widehat{g}_{\mytop})} \\
        &=\frac{\mathcal{R}(\overline{g}_{\mytop})-\mathcal{R}(\widehat{g}_{\mytop}) + \gamma \left(\overline{g}_{\mytop}\mathcal{R}(\widehat{g}_{\mytop}) - \widehat{g}_{\mytop}\mathcal{R}(\overline{g}_{\mytop})\right)}{(1- \gamma \overline{g}_{\mytop})(1- \gamma \widehat{g}_{\mytop})} \\
        &=\frac{\mathcal{R}(\overline{g}_{\mytop})-\mathcal{R}(\widehat{g}_{\mytop}) + \gamma \left(\overline{g}_{\mytop}- \widehat{g}_{\mytop}\right)\mathcal{R}(\widehat{g}_{\mytop}) + \gamma \widehat{g}_{\mytop}\left(\mathcal{R}(\widehat{g}_{\mytop}) -\mathcal{R}(\overline{g}_{\mytop})\right)}{(1- \gamma \overline{g}_{\mytop})(1- \gamma \widehat{g}_{\mytop})} \\
        &=\frac{(1-\gamma \widehat{g}_{\mytop})\left(\mathcal{R}(\widehat{g}_{\mytop}) -\mathcal{R}(\overline{g}_{\mytop})\right) + \gamma \left(\overline{g}_{\mytop}- \widehat{g}_{\mytop}\right)\mathcal{R}(\widehat{g}_{\mytop})}{(1- \gamma \overline{g}_{\mytop})(1- \gamma \widehat{g}_{\mytop})} \\
        &\leq \frac{(1-\gamma \widehat{g}_{\mytop})\alpha_{\mytop}\left(\overline{g}_{\mytop}- \widehat{g}_{\mytop}\right) + \gamma \left(\overline{g}_{\mytop}- \widehat{g}_{\mytop}\right)\mathcal{R}(\widehat{g}_{\mytop})}{(1- \gamma \overline{g}_{\mytop})(1- \gamma \widehat{g}_{\mytop})} \\
        &= \frac{(1-\gamma \widehat{g}_{\mytop})\alpha_{\mytop} + \gamma \mathcal{R}(\widehat{g}_{\mytop})}{(1- \gamma \overline{g}_{\mytop})(1- \gamma \widehat{g}_{\mytop})}\left(\overline{g}_{\mytop}- \widehat{g}_{\mytop}\right)
    \end{align}
    
    Reinjecting in Equation \ref{eq:max}, we have the following upper bound for $V^{\mydiese}_r(x) - \widehat{V}^{\mydiese}_r(x)$:
    \begin{align}
        x \max \left\{0,\frac{(1-\gamma \widehat{g}_{\mytop})\alpha_{\mytop} + \gamma \mathcal{R}(\widehat{g}_{\mytop})}{(1- \gamma \overline{g}_{\mytop})(1- \gamma \widehat{g}_{\mytop})}\left(\overline{g}_{\mytop}- \widehat{g}_{\mytop}\right),\frac{(1-\gamma \widehat{g}_{\mybot})\alpha_{\mybot} + \gamma \mathcal{R}(\widehat{g}_{\mybot})}{(1- \gamma \overline{g}_{\mybot})(1- \gamma \widehat{g}_{\mybot})}\left(\overline{g}_{\mybot}- \widehat{g}_{\mybot}\right)\right\}.
    \end{align}
    
    We observe that it is linear with $\overline{g}_{\mytop} - \widehat{g}_{\mytop}$, but with a constant that is not really one, since it depends on both $\overline{g}_{\mytop/\mybot}$ and $\widehat{g}_{\mytop/\mybot}$. We can further make replacement of either $\overline{g}_{\mytop/\mybot}$ or $\widehat{g}_{\mytop/\mybot}$ with $\epsilon_{\mytop/\mybot} = |\overline{g}_{\mytop/\mybot}- \widehat{g}_{\mytop/\mybot}|$, and use the following Taylor expansion to prove that the expression remains linear in $\epsilon_{\mytop/\mybot}$:
    \begin{align}
        \frac{1}{1-\gamma(g-\epsilon)} = \frac{1}{1-\gamma g} + \frac{\gamma\epsilon}{(1-\gamma g)^2} + o\left(\epsilon^2\right),
    \end{align}
    which concludes the first part of the Lemma.
    
    \textbf{(II)} When $\max_{g\geq 1} \mathcal{R}(g) > 0$, we may choose $\gamma$ such that $\max_{g\geq \frac{1}{\gamma}} \mathcal{R}(g)>0$. In this case, we can observe that the worst case scenario happens when $\alpha_{\mytop} > 0$\footnote{We could break down various cases to improve the constants of the bounds depending on each specific case, but we considered that it complicates the proof while the interesting part of the theorem is that the value decays linearly with the upper convex envelop domain divergence} and $\mathcal{R}(\overline{g}_{\mytop})- \mathcal{R}(\widehat{g}_{\mytop}) = \alpha_{\mytop}(\overline{g}_{\mytop} - \widehat{g}_{\mytop})$. This worst case scenario is easy to solve since Property \ref{prop:increasing} states that when $\max_{g\geq \frac{1}{\gamma}} \mathcal{R}(g)>0$, the optimal policy is decreasing with time and until reaching $x_{\mytop}$ when the optimal is to play $\argmax_{g\geq \frac{1}{\gamma}} \mathcal{R}(g)$, which in our worst case scenario equals $\overline{g}_{\mytop}$ in $M$ and $\widehat{g}_{\mytop}$ in $\widehat{M}$. We may conclude that $\forall x,$ $\pi^{\mydiese}(x) = \overline{g}_{\mytop}$ and $\widehat{\pi}^{\mydiese}(x) = \widehat{g}_{\mytop}$. With this information, we can compute the difference in value $V^{\mydiese}_r(x) - \widehat{V}^{\mydiese}_r(x)$:
    \begin{align}
        \sum_{t=0}^{t_{\mytop}-1} x (\gamma \overline{g}_{\mytop})^t \mathcal{R}(\overline{g}_{\mytop}) + \sum_{t=t_{\mytop}}^\infty x_{\mytop}\gamma^t \mathcal{R}(\overline{g}_{\mytop}) - \sum_{t=0}^{\widehat{t}_{\mytop}-1} x (\gamma \widehat{g}_{\mytop})^t \mathcal{R}(\widehat{g}_{\mytop}) - \sum_{t=\widehat{t}_{\mytop}}^\infty x_{\mytop}\gamma^t \mathcal{R}(\widehat{g}_{\mytop}),
    \end{align}
    where $t_{\mytop}$ (resp. $\widehat{t}_{\mytop}$)\footnote{For the sake of simplicity, we treat them as integer.} is the time to reach the maximal state $x_{\mytop}$:
    \begin{align}
        t_{\mytop} = \frac{\ln{\frac{x_{\mytop}}{x}}}{\ln{\overline{g}_{\mytop}}} \quad\quad\text{and}\quad\quad  \widehat{t}_{\mytop} = \frac{\ln{\frac{x_{\mytop}}{x}}}{\ln{\widehat{g}_{\mytop}}}. \label{eq:t-hat}
    \end{align}
    
    We proceed as follows to estimate $V^{\mydiese}_r(x) - \widehat{V}^{\mydiese}_r(x)$:
    \begin{align}
        &\leq \sum_{t=0}^{\widehat{t}_{\mytop}-1} x (\gamma \overline{g}_{\mytop})^t \mathcal{R}(\overline{g}_{\mytop}) + \sum_{t=\widehat{t}_{\mytop}}^\infty x_{\mytop}\gamma^t \mathcal{R}(\overline{g}_{\mytop}) - \sum_{t=0}^{\widehat{t}_{\mytop}-1} x (\gamma \widehat{g}_{\mytop})^t \mathcal{R}(\widehat{g}_{\mytop}) - \sum_{t=\widehat{t}_{\mytop}}^\infty x_{\mytop}\gamma^t \mathcal{R}(\widehat{g}_{\mytop}) \\
        &= \sum_{t=0}^{\widehat{t}_{\mytop}-1} x \gamma^t \left(\overline{g}_{\mytop}^t \mathcal{R}(\overline{g}_{\mytop}) - \widehat{g}_{\mytop}^t \mathcal{R}(\widehat{g}_{\mytop})\right) + \sum_{t=\widehat{t}_{\mytop}}^\infty x_{\mytop}\gamma^t \left(\mathcal{R}(\overline{g}_{\mytop}) - \mathcal{R}(\widehat{g}_{\mytop})\right) \\
        &= \sum_{t=0}^{\widehat{t}_{\mytop}-1} x \gamma^t \left(\overline{g}_{\mytop}^t \mathcal{R}(\overline{g}_{\mytop})+\widehat{g}_{\mytop}^t \mathcal{R}(\overline{g}_{\mytop})-\widehat{g}_{\mytop}^t \mathcal{R}(\overline{g}_{\mytop}) - \widehat{g}_{\mytop}^t \mathcal{R}(\widehat{g}_{\mytop})\right) + \frac{x_{\mytop}\alpha_{\mytop}\gamma^{\widehat{t}_{\mytop}}}{1-\gamma} \left(\overline{g}_{\mytop} - \widehat{g}_{\mytop}\right) \\
        &= \sum_{t=0}^{\widehat{t}_{\mytop}-1} x \gamma^t\mathcal{R}(\overline{g}_{\mytop}) \left(\overline{g}_{\mytop}^t - \widehat{g}_{\mytop}^t\right) +\sum_{t=0}^{\widehat{t}_{\mytop}-1} x (\gamma \widehat{g}_{\mytop})^t \left(\mathcal{R}(\overline{g}_{\mytop})-\mathcal{R}(\widehat{g}_{\mytop})\right) + \frac{x_{\mytop}\alpha_{\mytop}\gamma^{\widehat{t}_{\mytop}}}{1-\gamma} \left(\overline{g}_{\mytop} - \widehat{g}_{\mytop}\right) \\
        &= x \gamma\mathcal{R}(\overline{g}_{\mytop}) \sum_{t=0}^{\widehat{t}_{\mytop}-1} \gamma^{t-1} \left(\overline{g}_{\mytop}^t - \widehat{g}_{\mytop}^t\right)
        + x \frac{(\gamma \widehat{g}_{\mytop})^{\widehat{t}_{\mytop}}-1}{\gamma \widehat{g}_{\mytop}-1} \alpha_{\mytop}(\overline{g}_{\mytop} - \widehat{g}_{\mytop}) 
        + \frac{x_{\mytop}\alpha_{\mytop}\gamma^{\widehat{t}_{\mytop}}}{1-\gamma} \left(\overline{g}_{\mytop} - \widehat{g}_{\mytop}\right). \label{eq:paretodomainvalue}
    \end{align}
    
    The sum inside the first term requires a bit of work:
    \begin{align}
        \sum_{t=0}^{\widehat{t}_{\mytop}-1} \gamma^{t-1} \left(\overline{g}_{\mytop}^t - \widehat{g}_{\mytop}^t\right) &= \sum_{t=0}^{\widehat{t}_{\mytop}-1} \gamma^{t-1} \left(\overline{g}_{\mytop} - \widehat{g}_{\mytop}\right)\sum_{i=0}^{t-1} \overline{g}_{\mytop}^i \widehat{g}_{\mytop}^{t-1-i} \\
        &\leq \sum_{t=0}^{\widehat{t}_{\mytop}-1} \gamma^{t-1} \left(\overline{g}_{\mytop} - \widehat{g}_{\mytop}\right)\sum_{i=0}^{t-1} \overline{g}_{\mytop}^i \overline{g}_{\mytop}^{t-1-i} \\
        &= \left(\overline{g}_{\mytop} - \widehat{g}_{\mytop}\right)\sum_{t=0}^{\widehat{t}_{\mytop}-1} t(\gamma  \overline{g}_{\mytop})^{t-1} \\
        &= \left(\overline{g}_{\mytop} - \widehat{g}_{\mytop}\right) \sum_{t=0}^{\widehat{t}_{\mytop}-1}\frac{\partial\left((\gamma  \overline{g}_{\mytop})^t\right)}{\partial(\gamma  \overline{g}_{\mytop})} \\
        &= \left(\overline{g}_{\mytop} - \widehat{g}_{\mytop}\right) \frac{\partial\left(\sum_{t=0}^{\widehat{t}_{\mytop}-1}(\gamma  \overline{g}_{\mytop})^t\right)}{\partial(\gamma  \overline{g}_{\mytop})} \\
        &= \left(\overline{g}_{\mytop} - \widehat{g}_{\mytop}\right) \frac{\partial\left(\frac{(\gamma  \overline{g}_{\mytop})^{\widehat{t}_{\mytop}}-1}{\gamma  \overline{g}_{\mytop}-1}\right)}{\partial(\gamma  \overline{g}_{\mytop})} \\
        &= \left(\overline{g}_{\mytop} - \widehat{g}_{\mytop}\right) \frac{\left(\gamma  \overline{g}_{\mytop}\widehat{t}_{\mytop} - \gamma  \overline{g}_{\mytop} - \widehat{t}_{\mytop}\right)(\gamma  \overline{g}_{\mytop})^{\widehat{t}_{\mytop}-1}-1}{(\gamma  \overline{g}_{\mytop}-1)^2} \\
        &\leq \left(\overline{g}_{\mytop} - \widehat{g}_{\mytop}\right) \frac{\widehat{t}_{\mytop}(\gamma  \overline{g}_{\mytop})^{\widehat{t}_{\mytop}}}{(\gamma  \overline{g}_{\mytop}-1)^2},
    \end{align}
    which allows us to conclude the second part of the proof by reinjecting this expression into the optimal value error (Equation \ref{eq:paretodomainvalue}). Also, we may notice that $\widehat{g}_{\mytop}^{\widehat{t}_{\mytop}} = \frac{x_{\mytop}}{x}$:
    \begin{align}
        V^{\mydiese}_r(x) - \widehat{V}^{\mydiese}_r(x) &\leq \left(x \gamma\mathcal{R}(\overline{g}_{\mytop}) \frac{\widehat{t}_{\mytop}(\gamma  \overline{g}_{\mytop})^{\widehat{t}_{\mytop}}}{(\gamma  \overline{g}_{\mytop}-1)^2}
        + x \frac{\gamma^{\widehat{t}_{\mytop}}\frac{x_{\mytop}}{x}-1}{\gamma \widehat{g}_{\mytop}-1} \alpha_{\mytop}
        + \frac{x_{\mytop}\alpha_{\mytop}\gamma^{\widehat{t}_{\mytop}}}{1-\gamma}\right) \left(\overline{g}_{\mytop} - \widehat{g}_{\mytop}\right). \label{eq:lemma6-valuediff}
    \end{align}
    
    We observe that it is linear with $\overline{g}_{\mytop} - \widehat{g}_{\mytop}$, but with a constant that is not really one, since it depends on both $\overline{g}_{\mytop}$ and $\widehat{g}_{\mytop}$. We can still further make replacement of either $\overline{g}_{\mytop}$ or $\widehat{g}_{\mytop}$ with $\epsilon = |\overline{g}_{\mytop}- \widehat{g}_{\mytop}|$, and use the following Taylor expansion to prove that the expression remains linear in $\epsilon$:
    \begin{align}
        \frac{1}{\gamma \widehat{g}_{\mytop}-1} =  \frac{1}{\gamma(\overline{g}_{\mytop}-\epsilon)-1} = \frac{1}{\gamma \overline{g}_{\mytop}-1} + \frac{\gamma\epsilon}{(\gamma \overline{g}_{\mytop} - 1)^2} + \mathcal{O}\left(\epsilon^2\right).
    \end{align}
    
    Only $\widehat{t}_{\mytop}$ remains an uncontrolled variable for the moment. From \ref{eq:t-hat}, it is direct that:
    \begin{align}
        \widehat{t}_{\mytop} = \frac{\ln{\overline{g}_{\mytop}}}{\ln{\widehat{g}_{\mytop}}}t_{\mytop} &= t_{\mytop}\left(1+\frac{\ln{\frac{\overline{g}_{\mytop}}{\widehat{g}_{\mytop}}}}{\ln{\widehat{g}_{\mytop}}}\right)\\
        &= t_{\mytop}\left(1+\frac{\ln\left(1+\frac{\epsilon}{\widehat{g}_{\mytop}}\right)}{\ln{\widehat{g}_{\mytop}}}\right)\\
        &\leq t_{\mytop}\left(1+\frac{\epsilon}{\widehat{g}_{\mytop}\ln{\widehat{g}_{\mytop}}}\right).\\
        &\leq t_{\mytop}\left(1+\frac{\epsilon}{\overline{g}_{\mytop}\ln{\overline{g}_{\mytop}}} + \mathcal{O}(\epsilon^2)\right).
    \end{align}
    
    Injecting it back to Equation \ref{eq:lemma6-valuediff}, we obtain for $\frac{1}{\epsilon}\left(V^{\mydiese}_r(x) - \widehat{V}^{\mydiese}_r(x)\right)$:
    \begin{align}
        &\leq x \gamma\mathcal{R}(\overline{g}_{\mytop}) \frac{t_{\mytop}\left(1+\frac{\epsilon}{\overline{g}_{\mytop}\ln{\overline{g}_{\mytop}}}+\mathcal{O}(\epsilon^2)\right)(\gamma  \overline{g}_{\mytop})^{t_{\mytop}}(\gamma  \overline{g}_{\mytop})^{\frac{\epsilon t_{\mytop}}{\overline{g}_{\mytop}\ln{\overline{g}_{\mytop}}}+\mathcal{O}(\epsilon^2)}}{(\gamma  \overline{g}_{\mytop}-1)^2} + x \frac{\gamma^{t_{\mytop}}\frac{x_{\mytop}}{x}-1}{\gamma \overline{g}_{\mytop}-1} \alpha_{\mytop}
        + \frac{x_{\mytop}\alpha_{\mytop}\gamma^{t_{\mytop}}}{1-\gamma}+\mathcal{O}(\epsilon) \\
        &\leq x \gamma\mathcal{R}(\overline{g}_{\mytop}) \frac{t_{\mytop}\left(1+\mathcal{O}(\epsilon)\right)(\gamma  \overline{g}_{\mytop})^{t_{\mytop}}\left(1+\frac{\epsilon t_{\mytop}\ln\left(\gamma  \overline{g}_{\mytop}\right)}{\overline{g}_{\mytop}\ln{\overline{g}_{\mytop}}} + \mathcal{O}(\epsilon^2)\right)}{(\gamma  \overline{g}_{\mytop}-1)^2} + x \frac{\gamma^{t_{\mytop}}\frac{x_{\mytop}}{x}-1}{\gamma \overline{g}_{\mytop}-1} \alpha_{\mytop}
        + \frac{x_{\mytop}\alpha_{\mytop}\gamma^{t_{\mytop}}}{1-\gamma} +\mathcal{O}(\epsilon) \\
        &\leq x \gamma\mathcal{R}(\overline{g}_{\mytop}) \frac{t_{\mytop}(\gamma  \overline{g}_{\mytop})^{t_{\mytop}}\left(1+\frac{\epsilon t_{\mytop}}{\overline{g}_{\mytop}} + \mathcal{O}(\epsilon^2)\right)}{(\gamma  \overline{g}_{\mytop}-1)^2}
        + x \frac{\gamma^{t_{\mytop}}\frac{x_{\mytop}}{x}-1}{\gamma \overline{g}_{\mytop}-1} \alpha_{\mytop}
        + \frac{x_{\mytop}\alpha_{\mytop}\gamma^{t_{\mytop}}}{1-\gamma} +\mathcal{O}(\epsilon)\\
        &\leq x \gamma\mathcal{R}(\overline{g}_{\mytop}) \frac{t_{\mytop}(\gamma  \overline{g}_{\mytop})^{t_{\mytop}}}{(\gamma  \overline{g}_{\mytop}-1)^2}
        + x \frac{\gamma^{t_{\mytop}}\frac{x_{\mytop}}{x}-1}{\gamma \overline{g}_{\mytop}-1} \alpha_{\mytop}
        + \frac{x_{\mytop}\alpha_{\mytop}\gamma^{t_{\mytop}}}{1-\gamma} +\mathcal{O}(\epsilon),
    \end{align}
    which concludes the second part of the Lemma.
\end{proof}

\begin{lemma}[OeMDP model error]
    Let $M$ and $M'$ be two OeMDPs induced by $K$ arms of respective parameters $\{(\overline{r}_k,\overline{g}_k)\}_{k\in[K]}$ and $\{(\overline{r}'_k,\overline{g}'_k)\}_{k\in[K]}$. Then, for any optimal policy $\psi$ in $M$, we have the following upper bound on the value error $V^\psi_M(x) - V^\psi_{M'}(x)$ that decreases linearly with their model distance: $\max_{k\in[K]} |\overline{r}_k -\overline{r}'_k|$ and $\max_{k\in[K]} |\overline{g}_k -\overline{g}'_k|$.
    \label{lem:OeMDPmodelerror}
\end{lemma}
\begin{proof}
    This proof is very similar to that of Lemma \ref{lem:domainpareto}. We split the proof in two cases: \textbf{(I)} when $\forall k\in[K]$, such that $\overline{g}_k \geq 1$, $\overline{r}_k \leq 0$, and \textbf{(II)} otherwise, when $\exists k\in[K]$, such that $\overline{g}_k \geq 1$ and $\overline{r}_k > 0$. 
    
    \textbf{(I)} When $\forall k\in[K]$, such that $\overline{g}_k \geq 1$, $\overline{r}_k \leq 0$, we know that the optimal policy is constant for all $x$:
    \begin{align}
        \psi(x) = k^* \in \argmax_{k \in [K] \text{ s.t. } \overline{g}_k < 1} \frac{\overline{r}_k}{1- \gamma \overline{g}_k}.
    \end{align}
    
    Then, under the assumption that $\overline{g}'_{k^*} \leq 1$, which is mild since $\overline{g}_{k^*} < 1$, the error in value is direct:
    \begin{align}
        V^\psi_M(x) - V^\psi_{M'}(x) &= \frac{\overline{r}_{k^*}}{1- \gamma \overline{g}_{k^*}} - \frac{\overline{r}'_{k^*}}{1- \gamma \overline{g}'_{k^*}} \\
        &= \frac{\overline{r}_{k^*}-\overline{r}'_{k^*} + \gamma \left(\overline{g}_{k^*}- \overline{g}'_{k^*}\right)\overline{r}'_{k^*} + \gamma \overline{g}'_{k^*}\left(\overline{r}'_{k^*} -\overline{r}_{k^*}\right)}{(1- \gamma \overline{g}_{k^*})(1- \gamma \overline{g}'_{k^*})} \\
        &= \frac{\overline{r}_{k^*}-\overline{r}'_{k^*}}{1- \gamma \overline{g}_{k^*}} + \gamma\overline{r}'_{k^*}\frac{ \overline{g}_{k^*}- \overline{g}'_{k^*} }{(1- \gamma \overline{g}_{k^*})(1- \gamma \overline{g}'_{k^*})}\\
        &\leq \frac{\max_{k\in[K]} |\overline{r}_k -\overline{r}'_k|}{1- \gamma \overline{g}_{k^*}} + \gamma\overline{r}'_{k^*}\frac{\max_{k\in[K]} |\overline{g}_k -\overline{g}'_k|}{(1- \gamma \overline{g}_{k^*})(1- \gamma \overline{g}'_{k^*})}
    \end{align}
    which concludes the first part of the proof.
    
    \textbf{(II)} When $\exists k\in[K]$, such that $\overline{g}_k \geq 1$ and $\overline{r}_k > 0$, we may choose $\gamma$ such that $\exists k\in[K]$, such that $\overline{g}_k \geq \frac{1}{\gamma}$ and $\overline{r}_k > 0$. In this case, we can observe that the worst case scenario happens when:
    \begin{align}
        k_* \in \argmax_{k \in [K]} \overline{g}_k \cap \argmax_{k \in [K]} \overline{r}_k \cap \argmax_{k \in [K]} (\overline{g}_k - \overline{g}'_k) \cap \argmax_{k \in [K]} (\overline{r}_k - \overline{r}'_k)
    \end{align}
    
    It is direct to notice that in this worst case scenario, the optimal policy also happens to be constant: $\forall x, \psi(x) = k_*$. From now on, the proof is identical to that of Lemma \ref{lem:domainpareto} \textbf{(II)} with the following result:
    \begin{align*}
        V^\psi_M(x) - V^\psi_{M'}(x) &\leq x \gamma\overline{r}'_{k^*} \frac{t_{\mytop}(\gamma  \overline{g}'_{k^*})^{t_{\mytop}}}{(\gamma  \overline{g}'_{k^*}-1)^2}\left(\overline{g}'_{k^*} - \overline{g}_{k^*}\right)
        + x \frac{\gamma^{t_{\mytop}}\frac{x_{\mytop}}{x}-1}{\gamma \overline{g}_{k^*}-1} \left(\overline{r}'_{k^*} - \overline{r}_{k^*}\right) + \frac{x_{\mytop}\gamma^{t_{\mytop}}}{1-\gamma}  \left(\overline{r}'_{k^*} - \overline{r}_{k^*}\right), \text{ with } t_{\mytop} = \frac{\ln{\frac{x_{\mytop}}{x}}}{\ln{\overline{g}_{k^*}}} 
    \end{align*}
\end{proof}

\banditthm*
\begin{proof}
    Cases (a-b) result is directly stems from Lemma \ref{lem:banditcaseab}, where only the dependencies in $K$, $\delta$, and $T$ are retained.
    
    Case (c) result, first part, is proven in Lemma \ref{lem:banditcasec1}.
    
    And finally, Case (c) result, second part, is demonstrated in Lemma \ref{lem:banditcasec2}.
\end{proof}

\begin{lemma}
    If the problem is Case (a-b), then the regret of Algorithm \ref{alg:main} is $\text{Regret}(K, \delta, T,\epsilon) \in \mathcal{O}\left(\frac{K\ln\frac{1}{\delta}}{\epsilon}+K\delta T\right)$, as a function of $K$ the number of arms, $\delta$ a concentration probability hyperparameter for the algorithm, $T$ the total number of pulls, and $\epsilon$ the decidability of the setting.
    \label{lem:banditcaseab}
\end{lemma}
\begin{proof}
    \textbf{Disclaimer:} The proof of this lemma is kept as a sketch for the sake of simplicity.
    
    \begin{figure}[h]
        \centering
        \begin{tikzpicture}[every node/.append style={font=\scriptsize}]
        
            \fill[yellow!50!white,opacity=0.5] (0,0) rectangle (4,4);
            \fill[orange!20!white,opacity=0.5] (2,2) rectangle (6,6);
            \draw [->,orange] (1.4,2.8) -- (3.35,4.75);
            \draw [->,green] (2,2) -- (5.95,5.95);
            \fill [blue] (2,2) circle(2pt);
            \draw(2,1.8) node{$(\overline{g}_k,\overline{r}_k)$};
            \fill [yellow] (1.4,2.8) circle(2pt);
            \draw(1.4,2.6) node{$(\hat{g}_k,\hat{r}_k)$};
            \draw(2.4,3.8) node{\tiny{$+\frac{\xi}{\sqrt{n_k}}$}};
            \fill [orange] (3.4,4.8) circle(2pt);
            \draw(3.4,5.1) node{$(g^{\myplus}_k,r^{\myplus}_k)$};
            \fill [green] (6,6) circle(2pt);
            \draw(4,4) node{\tiny{$+\frac{2\xi}{\sqrt{n_k}}$}};
            \draw(6,6.3) node{$(g^{\dag}_k,r^{\dag}_k)$};
        \end{tikzpicture}
        \caption{The true parameters (blue mark) are estimated with the empirical mean (yellow mark). This estimate is within the yellow area with high probability $1-\delta$. The optimistic estimate (orange mark) is shifted by $\frac{\xi}{\sqrt{n_k}}$ and is therefore within the orange area with high probability $1-\delta$. $(g^{\dag}_k,r^{\dag}_k)$ therefore constitutes the most defavorable outcome for the optimistic estimate to infer that the setting is in Case (a-b).}
        \label{fig:thalgo1}
    \end{figure}
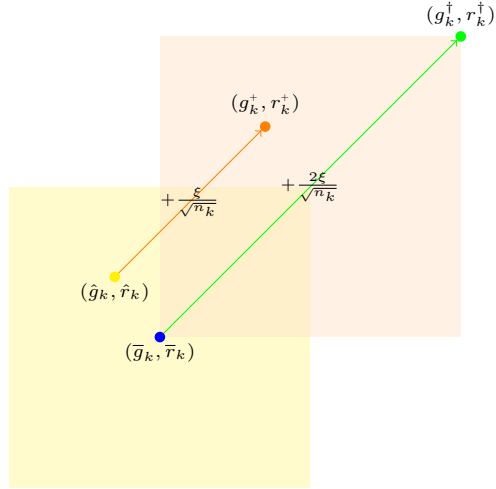
    We start by expressing the optimistic parameters for the reward and the growth of arms (in orange on Figure \ref{fig:thalgo1}):
    \begin{align}
        r^{\myplus}_k = \hat{r}_k + \frac{\xi}{\sqrt{n_k}}\quad\text{and}\quad  g^{\myplus}_k = \hat{g}_k + \frac{\xi}{\sqrt{n_k}},
    \end{align}
    where $\hat{r}_k$ and $\hat{g}_k$ are the empirical means (in yellow) of, respectively, reward and growth for Arm $k$, and where $n_k$ is the number of times Arm $k$ has been pulled. For each arm $k$, let $e_k$ be the maximal difference between the true parameters $(\overline{g}_k,\overline{r}_k)$ (in blue) and their empirical means $(\hat{g}_k,\hat{r}_k)$ with high probability $1-\delta$, obtained thanks to 2-sided Hoeffding:
    \begin{align}
        \hat{g}_k &\in [\overline{g}_k-e_k,\overline{g}_k+e_k] \quad\text{and}\quad \hat{r}_k \in [\overline{r}_k-e_k,\overline{r}_k+e_k] \\
        \text{with}\quad e_k &= \frac{\max(\dot{g}_{\mytop},\dot{r}_{\mytop}-\dot{r}_{\mybot})\sqrt{\ln\frac{2}{\delta}}}{\sqrt{2n_k}}.
    \end{align} 
    
    The empirical estimate has therefore to be in the yellow area with high probability $1-\delta$. If we choose $e_k=\frac{\xi}{\sqrt{n_k}}$, and therefore:
    \begin{align}
        \xi = \frac{\max(\dot{g}_{\mytop},\dot{r}_{\mytop}-\dot{r}_{\mybot})\sqrt{\ln\frac{2}{\delta}}}{\sqrt{2}},
    \end{align}
    then we have:
    \begin{align}
        g^{\myplus}_k\in[\overline{g}_k,\overline{g}_k+2e_k]\quad\text{and}\quad g^{\myplus}_k\in[\overline{r}_k,\overline{r}_k+2e_k],
    \end{align}
    \textit{i.e.} the optimistic estimate has to be in the orange area with high probability $1-\delta$. Graphically, we observe that the worst situation happens when the optimistic parameters hit their upper bound: $g^{\dag}_k =  \overline{g}_k+2e_k$ and $r^{\dag}_k = \overline{r}_k+2e_k$ (in green). 
    
    The algorithm will only play arms that are on the optimistic convex hull, the orange broken line on Figure \ref{fig:thalgo2}. With time, it will take the form of a line, that is pushed down-left, as more pulls are performed. Once, it is pushed down below the critical point (1,0) (or more rigorously the red semi-line), the case is identified as being Case (a-b), and then the corresponding strategy is applied. The orange line depends on the random pulls outcomes and it is more convenient to consider the green line which dominates the orange line with high probability $1-K\delta$, and we are going to measure the regret until getting the green line under the critical point (1,0).
    \begin{figure}[h]
        \centering
        \def\trueposition{{{1.75,1.1}, {2.8,-0.7}, {3.1,-0.2}, {4.5,-1.1}, {4.3, -2.5}, {9.1,-5.5}}}
        \def\empiricposition{{{0.5,1.5}, {2.4,-1}, {3,0.3}, {5,-1.2}, {5.2, -2.3}, {8,-5}}}
        \def\optimisticposition{{{2.5,3.5}, {4.8,1.4}, {4.5,1.8}, {6.3,0.1}, {6.8, -0.7}, {9.8,-3.2}}}
         \begin{tikzpicture}[every node/.append style={font=\scriptsize}]
        
            \draw [->,black] (0,-6) -- (13,-6) ;
            \draw [->,black] (0,-6) -- (0,5.5) ;
            \draw [-,red] (4,0) -- (13,0) ;
            \fill [red] (4,0) circle(2pt);
            \draw [red] (4,-0.25) node{$(1,0)$};
            \foreach \i in {0,...,5}{
                \def\j{\i}
                \draw [densely dotted,->,orange] (\empiricposition[\i][0],\empiricposition[\i][1]) -- (\optimisticposition[\i][0]-0.05,\optimisticposition[\i][1]-0.05);
                \draw [densely dotted,->,green] (\trueposition[\i][0],\trueposition[\i][1]) -- (\trueposition[\i][0]+2*\optimisticposition[\i][0]-2*\empiricposition[\i][0]-0.05,\trueposition[\i][1]+2*\optimisticposition[\i][1]-2*\empiricposition[\i][1]-0.05);
                \fill [blue] (\trueposition[\i][0],\trueposition[\i][1]) circle(2pt);
                \draw [blue] (\trueposition[\i][0],\trueposition[\i][1]-0.2) node{\j};
                \fill [yellow] (\empiricposition[\i][0],\empiricposition[\i][1]) circle(2pt);
                \draw [yellow] (\empiricposition[\i][0],\empiricposition[\i][1]-0.2) node{\j};
                \fill [orange] (\optimisticposition[\i][0],\optimisticposition[\i][1]) circle(2pt);
                \draw [orange] (\optimisticposition[\i][0],\optimisticposition[\i][1]+0.2) node{\j};
                \fill [green] (\trueposition[\i][0]+2*\optimisticposition[\i][0]-2*\empiricposition[\i][0],\trueposition[\i][1]+2*\optimisticposition[\i][1]-2*\empiricposition[\i][1]) circle(2pt);
                \draw [green] (\trueposition[\i][0]+2*\optimisticposition[\i][0]-2*\empiricposition[\i][0],\trueposition[\i][1]+2*\optimisticposition[\i][1]-2*\empiricposition[\i][1]+0.2) node{\j};
            }
            \draw [-,blue] (1.75,1.1) -- (4.5,-1.1) ;
            \draw [-,blue] (4.5,-1.1) -- (9.1,-5.5) ;
            \draw [-,green] (5.75,5.1) -- (7.6,4.1) ;
            \draw [-,green] (7.6,4.1) -- (12.7,-1.9) ;
            \draw [-,orange] (2.5,3.5) -- (4.5,1.8) ;
            \draw [-,orange] (4.5,1.8) -- (6.3,0.1) ;
            \draw [-,orange] (6.3,0.1) -- (9.8,-3.2) ;
            
        
        
        \end{tikzpicture}
        \caption{We keep the same colour code as for Figure \ref{fig:thalgo1}: the true parameters are in blue, the empirical mean in yellow, the optimistic estimate in orange and the upper bound of the optimistic estimate in green. Additionally, the critical semi-line is shown in red. Since the green Pareto front always dominates the orange one, and since the orange Pareto position with respect to the critical semi-line determines the decision apply Case (a-b) policy, we may show that this decision will be taken at the latest when the green Pareto goes under the critical semi-line.}
        \label{fig:thalgo2}
    \end{figure}
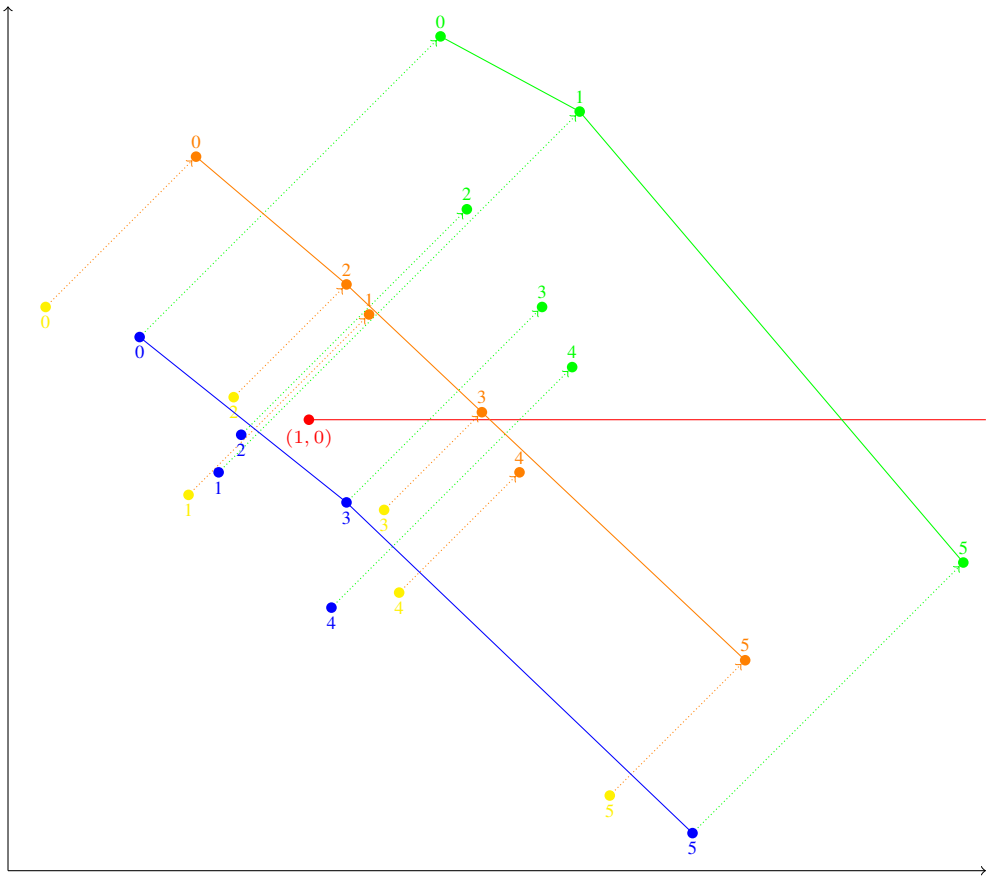
    
    When that happens, the slope of the orange line depends on the parameters of the setting and may also differ from one run to another, but in any configuration (see Remark \ref{rm:discussionpulls} for more), if $\epsilon_k$ is the $\ell_\infty$ distance between point $(\overline{g}_k,\overline{r}_k)$ and the straight orange line that the optimist estimates converge to, with high probability $1-K\delta$, the total number of pulls on each arm is of the order of $\frac{\xi^2}{\epsilon_k^2}$, which occurs a total regret in the order of $\sum_k \frac{\xi^2}{\epsilon_k} \leq \frac{K\xi^2}{\epsilon}$ before starting the crowd decrease. The crowd decrease induces a constant regret close enough to $V^*_o(x_{\mytop})-V^*_o(x_0)$, which is positive in Case (b) (so, to be substracted from the regret), and negative in Case (a) (so, to be added to the regret). Please also note that, in Case (a), it is probable that $\epsilon_k$ is large for every Arm $k$ and that the maximum is never to be reached. With complementary probability $K\delta$, nothing can be said about the algorithm expect that the regret is linear with $T$ (the regret is smaller on expectation that the expected reward of the worst arm times the number of pulls). The total regret is therefore of order:
    \begin{align}
        \text{Regret}(K, \delta, T,\epsilon) \in \mathcal{O}\left(\frac{K\ln\frac{1}{\delta}}{\epsilon}+K\delta T\right).
    \end{align}
    
    \begin{remark}[Discussion around an upper bound on the number of pulls]
        In the end, the process amounts to having sufficient precision to make the straight line formed with the optimistic estimates go below critical point (1,0) for a pair of arms, and make sure that all optimistic estimates of other arms are below that straight line. It also has to be noted that, once the maximal crowd reached, the chosen interpolation between arms is 1 (or it means that the optimistic estimate of the arms considers the maximal reward arm to have a growth larger than 1, which means that it will be pulled deterministically until this is not the case anymore), which implies an actual expected growth lower than one, but then the next chosen interpolation will ultimately compensate for the crowd loss by choosing a higher crowd. Over time, the overall growth will be at least the one aimed at: 1. The amount of pulls may get significantly larger if the orange line converges very close to one specific arm, because it means that this arm would need to be pulled many times to be sufficiently precise. Further, we study from a single pair of arms how many pulls are required to reach a sufficient precision. For simplicity, let us assume that these two arms are $k=1,2$. We know that the straight line $(d_{1,2})$ passing through $(\overline{g}_1,\overline{r}_1)$ and $(\overline{g}_2,\overline{r}_2)$ is at distance $\epsilon_{1,2}$ below the critical point $(1,0)$, which gives us that:
        \begin{align}
            (d_{1,2}):\quad y&=\frac{\overline{r}_2-\overline{r}_1}{\overline{g}_2-\overline{g}_1}x + \frac{\overline{r}_1 \overline{g}_2-\overline{r}_2 \overline{g}_1}{\overline{g}_2-\overline{g}_1} \\
            \overline{r}_2-\overline{r}_1 + \overline{r}_1 \overline{g}_2-\overline{r}_2 \overline{g}_1 &= -\epsilon_{1,2} \sqrt{(\overline{r}_2-\overline{r}_1)^2+(\overline{g}_2-\overline{g}_1)^2} \\
            \overline{r}_2(1-\overline{g}_1) + \overline{r}_1(\overline{g}_2-1) &= -\epsilon_{1,2} \sqrt{(\overline{r}_2-\overline{r}_1)^2+(\overline{g}_2-\overline{g}_1)^2} \coloneqq \rho_{1,2}
        \end{align}
        
        We are interested in finding the conditions for the straight line $(d^{\dag}_{1,2})$ passing through $(g^{\dag}_1,r^{\dag}_1)$ and $(g^{\dag}_2,r^{\dag}_2)$ to be below the critical point $(1,0)$, which gives:
        \begin{align}
            (d^{\dag}_{1,2}):\quad y =\frac{r^{\dag}_2-r^{\dag}_1}{g^{\dag}_2-g^{\dag}_1}x + \frac{r^{\dag}_1 g^{\dag}_2-r^{\dag}_2 g^{\dag}_1}{g^{\dag}_2-g^{\dag}_1}& \\
            r^{\dag}_2(1-g^{\dag}_1) + r^{\dag}_1(g^{\dag}_2-1) &\leq 0 \\
            \left(\overline{r}_2 + \frac{2\xi}{\sqrt{n_2}}\right)\left(1-\overline{g}_1-\frac{2\xi}{\sqrt{n_1}}\right) + \left(\overline{r}_1 + \frac{2\xi}{\sqrt{n_1}}\right)\left(\overline{g}_2 + \frac{2\xi}{\sqrt{n_2}}-1\right) &\leq 0 \\
            \rho_{1,2} \leq \frac{2\xi}{\sqrt{n_2}}(1-\overline{g}_1 + \overline{r}_1) + \frac{2\xi}{\sqrt{n_1}}(\overline{g}_2-1-\overline{r}_2).&
        \end{align}
        
        However, $n_1$ and $n_2$ are related: 
        \begin{align}
            n_1 \approx \alpha_{1,2} N_{1,2} \quad\text{and}\quad n_2 &\approx (1-\alpha_{1,2})N_{1,2} \quad\text{with}\quad\alpha_{1,2}=\frac{\overline{g}_2-1}{\overline{g}_2-\overline{g}_1}. \\
            n_1 \approx \frac{\alpha_{1,2}}{1-\alpha_{1,2}}n_2 &= \frac{\overline{g}_2-1}{1-\overline{g}_1}n_2
        \end{align}
        
        Injecting this, we get that having:
        \begin{align}
            \sqrt{N_{1,2}} &\geq \frac{2\xi}{\rho_{1,2}}\left(\frac{\overline{g}_2-1-\overline{r}_2}{\sqrt{\alpha_{1,2}}}+\frac{1-\overline{g}_1+\overline{r}_1}{\sqrt{1-\alpha_{1,2}}}\right) \\
            &\geq \frac{2\xi}{\rho_{1,2}}\left(\frac{\alpha_{1,2}(\overline{g}_2-\overline{g}_1)-\overline{r}_2}{\sqrt{\alpha_{1,2}}}+\frac{(1-\alpha_{1,2})(\overline{g}_2-\overline{g}_1)+\overline{r}_1}{\sqrt{1-\alpha_{1,2}}}\right) \\
            &\geq \frac{2\xi}{\rho_{1,2}}\left(\sqrt{\alpha_{1,2}}(\overline{g}_2-\overline{g}_1) - \frac{\overline{r}_2}{\sqrt{\alpha_{1,2}}} + \sqrt{1-\alpha_{1,2}}(\overline{g}_2-\overline{g}_1) + \frac{\overline{r}_1}{\sqrt{1-\alpha_{1,2}}}\right)\coloneqq T_{1,2}
        \end{align}
        guarantees that the optimistic convex envelop is below the critical semi-line $[d_0)$. Moreover:
        \begin{align}
          T_{1,2} &= 2\xi \left(\left(\sqrt{\alpha_{1,2}}+\sqrt{1-\alpha_{1,2}}\right)\frac{\overline{g}_2-\overline{g}_1}{\rho_{1,2}} - \frac{\overline{r}_2}{\sqrt{\alpha_{1,2}}} + \frac{\overline{r}_1}{\sqrt{1-\alpha_{1,2}}}\right)\\
          &\leq \frac{2\xi}{\epsilon_{1,2}} \left(\sqrt{2} + \frac{\overline{r}_1 - \overline{r}_2}{\sqrt{(\overline{r}_2-\overline{r}_1)^2+(\overline{g}_2-\overline{g}_1)^2}}\left(\frac{1}{\sqrt{\alpha_{1,2}}} + \frac{1}{\sqrt{1-\alpha_{1,2}}}\right)\right)\\
          &\leq \frac{2\xi}{\epsilon_{1,2}} \left(\sqrt{2} + \frac{1}{\sqrt{\alpha_{1,2}}} + \frac{1}{\sqrt{1-\alpha_{1,2}}}\right)
        \end{align}
        
        We observe that $N_{1,2}$ might get large when either $\overline{g}_1$ ($\alpha$ is close to 1) or $\overline{g}_2$ ($\alpha$ is close to 0) are close to 1. If $\overline{g}_1$ is very close to 1, then it means that $r^{\dag}_1$ will soon be negative and Arm 1 will not be selected anymore. If $\overline{g}_2$ is very close to 1, then after selecting Arm $2$ often times, a pair of $k=1$ and some $k=3$ such that $\overline{g}_3$ should be larger than $\overline{g}_2$. The worst case would consist of $k=2$ being the only arm with $\overline{g}_2>1$ (and still $\overline{g}_2$ very close to 1). 
        \label{rm:discussionpulls} 
    \end{remark}
\end{proof}

\begin{lemma}
    If there exists $k_*$, such that $\overline{g}_* \geq 1$ and $\overline{r}_* = \max_k \overline{r}_k \geq 0$, the regret of Algorithm \ref{alg:main} is:
    \begin{align}
        \text{Regret}(K, \delta, T,\epsilon) \in \mathcal{O}\left(K\delta T + K\ln\frac{1}{\delta}\right),
    \end{align}
    as a function of $K$ the number of arms, $\delta$ a concentration probability hyperparameter for the algorithm, $T$ the total number of pulls, and $\epsilon$ the decidability of the setting.
    \label{lem:banditcasec1}
\end{lemma}
\begin{proof}
    We assume here, that there exists $k_*$, such that $\overline{g}_* \geq 1$ and $\overline{r}_* = \max_k \overline{r}_k \geq 0$. At each time step, Algorithm \ref{alg:main} plays an arm $k$ such that $r^{\myplus}_k \geq r^{\myplus}_*$. With high probability $1-K\delta$, we know that, for all $k$, $r^{\myplus}_k\in[\overline{r}_k,r^{\dag}_k]$. As a consequence, each arm $k\neq k_*$ may be pulled only if $r^{\dag}_k \geq \overline{r}_*$, which may happen a maximum $n_k$ times:
    \begin{align}
        n_k = \frac{(\dot{r}_{\mytop}-\dot{r}_{\mybot})^2\ln\frac{1}{\delta}}{\Delta_k^2},\quad\text{where }\Delta_k=\overline{r}_*-\overline{r}_k\text{ is the reward gap with Arm }k,
    \end{align}
    which yields an expected regret of:
    \begin{align}
        \Delta_k n_k = \frac{(\dot{r}_{\mytop}-\dot{r}_{\mybot})^2\ln\frac{1}{\delta}}{\Delta_k},
    \end{align}
    and therefore a total regret\footnote{We replace $\sum_{k\in[K]}\frac{1}{\Delta_k}$ with $K$ in the order of magnitude.} of:
    \begin{align}
        \mathcal{O}\left(K\ln\frac{1}{\delta}\right).
    \end{align}
    
    With complementary probability $K\delta$, we are in the concentration failure mode and we suffer a linear regret as a function of $T$. The overall regret is therefore:
    \begin{align}
        \mathcal{O}\left(K\delta T + K\ln\frac{1}{\delta}\right).
    \end{align}
\end{proof}

\begin{lemma}
    If the problem in Case (c), and there does not exist $k_*$, such that $\overline{g}_* \geq 1$ and $\overline{r}_* = \max_k \overline{r}_k \geq 0$, the regret of Algorithm \ref{alg:main} is:
    \begin{align}
        \text{Regret}(K, \delta, T,\epsilon) \in \mathcal{O}\left(K\delta T + K\ln\frac{1}{\delta} + \sqrt{T\ln\frac{1}{\delta}}\right),
    \end{align}
    as a function of $K$ the number of arms, $\delta$ a concentration probability hyperparameter for the algorithm, $T$ the total number of pulls, and $\epsilon$ the decidability of the setting.
    \label{lem:banditcasec2}
\end{lemma}
\begin{proof}
    We assume here, that there \underline{does not} exist $k_*$, such that $\overline{g}_* \geq 1$ and $\overline{r}_* = \max_k \overline{r}_k \geq 0$. It means that, there is either an optimal pair of arms $(k_1,k_2)$ that should be played with an interpolation parameter $\alpha=\frac{\overline{g}_2-1}{\overline{g}_2-\overline{g}_1}$. With high probability $1-K\delta$, we know that, for all $k$, $r^{\myplus}_k\in[\overline{r}_k,r^{\dag}_k]$. As a consequence, each arm $k\notin \{k_1,k_2\}$ may be pulled only if the point $(g^{\dag}_k, r^{\dag}_k)$ is over the line $(d_{1,2})$ passing through $(\overline{g}_1,\overline{r}_1)$ and $(\overline{g}_2,\overline{r}_2)$, which may happen a maximum $n_k$ times:
    \begin{align}
        n_k = \frac{\max(\dot{g}_{\mytop},\dot{r}_{\mytop}-\dot{r}_{\mybot})^2\ln\frac{1}{\delta}}{\Delta_k^2},\quad\text{where }\Delta_k\text{ is the distance of }(g^{\dag}_k, r^{\dag}_k) \text{ from }(d_{1,2}),
    \end{align}
    and therefore a total regret\footnote{We replace $\sum_{k\in[K]}\frac{1}{\Delta^2_k}$ with $K$ in the order of magnitude.} of:
    \begin{align}
        \mathcal{O}\left(K\ln\frac{1}{\delta}\right).
    \end{align}
    
    It may also happen that the wrong ratio $\hat{\alpha} > \alpha$ is used. This means that the played growth $\hat{g}$ is actually lower than one, and the regret at each time step is of order $1-\hat{g}$:
    \begin{align}
        \hat{g} &= \hat{\alpha}\overline{g}_1 + (1-\hat{\alpha})\overline{g}_2 \\
        &=\frac{g^{\myplus}_2-1}{g^{\myplus}_2-g^{\myplus}_1}\overline{g}_1 + \frac{1-g^{\myplus}_1}{g^{\myplus}_2-g^{\myplus}_1}\overline{g}_2 \\
        &=\frac{\overline{g}_2-\overline{g}_1 + \overline{g}_1 g^{\myplus}_2-g^{\myplus}_1 \overline{g}_2}{g^{\myplus}_2-g^{\myplus}_1} \\
        1-\hat{g} &= \frac{\overline{g}_1-\overline{g}_2 + (1-\overline{g}_1)g^{\myplus}_2-g^{\myplus}_1(\overline{g}_2-1)}{g^{\myplus}_2-g^{\myplus}_1}\\
        &\leq \frac{\overline{g}_1-\overline{g}_2 + (1-\overline{g}_1)(\overline{g}_2+2e_2)+(\overline{g}_1+2e_1)(\overline{g}_2-1)}{g^{\myplus}_2-g^{\myplus}_1} \quad \text{where } e_k = \frac{\max(\dot{g}_{\mytop},\dot{r}_{\mytop}-\dot{r}_{\mybot})\sqrt{\ln\frac{2}{\delta}}}{\sqrt{2n_k}} \\
        &= 2\frac{e_2(1-\overline{g}_1)+e_1(\overline{g}_2-1)}{g^{\myplus}_2-g^{\myplus}_1}.
    \end{align}
    
    We use here the same trick as in Lemma \ref{lem:banditcaseab}: since we know that Algorithm \ref{alg:main} will ultimately maintain crowd, we will experience an overall growth of 1, and therefore we know that $n_1$ and $n_2$ are tied together with the true ratio $\alpha$: $n_1\approx \alpha N_{1,2}$ and $n_2\approx (1-\alpha)N_{1,2}$. We therefore get:
    \begin{align}
        1-\hat{g} &\leq  \frac{\max(\dot{g}_{\mytop},\dot{r}_{\mytop}-\dot{r}_{\mybot})\sqrt{2\ln\frac{2}{\delta}}}{\sqrt{N_{1,2}}}\frac{\frac{1-\overline{g}_1}{\sqrt{\alpha}}+\frac{\overline{g}_2-1}{\sqrt{1-\alpha}}}{g^{\myplus}_2-g^{\myplus}_1} \\
        &\in \mathcal{O}\left(\sqrt{\frac{\ln\frac{1}{\delta}}{T}}\right).
    \end{align}
    
    If we sum over $T$ timesteps, we get a cumulative regret in $\mathcal{O}\left(\sqrt{T\ln\frac{1}{\delta}}\right)$
    
    With complementary probability $K\delta$, we are in the concentration failure mode and we suffer a linear regret as a function of $T$. The overall regret is therefore:
    \begin{align}
        \mathcal{O}\left(K\delta T + K\ln\frac{1}{\delta} + \sqrt{T\ln\frac{1}{\delta}}\right),
    \end{align}
    which concludes the proof.
\end{proof}

\end{document}